\documentclass{article}

\newcommand{\accuracyTable}{{
\begin{table*}[t]
\centering
\captionof{table}{Performance of LLaMA-7B compressed by \sysname, \sysname \texttt{(W)}, and baselines under different compression ratio (corresponding weight memory is listed inside bracket) on two language modeling datasets (measured by perplexity  (\textcolor{mygreen}{$\downarrow$})), eight common sense reasoning datasets (six measured by both individual and average accuracy (\textcolor{mygreen}{$\uparrow$}), TruthfulQA measured by BLEU score (\textcolor{mygreen}{$\uparrow$}), and GSM8K measured by Exact Match Accuracy (\textcolor{mygreen}{$\uparrow$})). The best performance is marked in bold. The relative performance gain compared to the best-performing baseline is marked in green inside bracket.
}
\vspace{-2mm}
\resizebox{1\textwidth}{!}{
\begin{tabular}{ccccccccccccc}
\midrule
\multicolumn{1}{c|}{\textsc{Ratio (Mem.)}}       & \multicolumn{1}{c|}{\textsc{Method}}    & WikiText-2\textcolor{mygreen}{$\downarrow$} & \multicolumn{1}{c|}{C4\textcolor{mygreen}{$\downarrow$}} & Openb. & ARC\_e & WinoG. & HellaS.  & PIQA & MathQA & \textbf{Average\textcolor{mygreen}{$\uparrow$}}  & TruthfulQA\textcolor{mygreen}{$\uparrow$} & GSM8K\textcolor{mygreen}{$\uparrow$}      \\ \midrule
\multicolumn{1}{c|}{\color[HTML]{9B9B9B}0\% (13.5 GB)}  & \multicolumn{1}{c|}{\color[HTML]{9B9B9B}Original}  & {\color[HTML]{9B9B9B}5.68}     & \multicolumn{1}{c|}{\color[HTML]{9B9B9B}7.34}      & {\color[HTML]{9B9B9B} 0.34} & {\color[HTML]{9B9B9B} 0.75} & {\color[HTML]{9B9B9B} 0.70} & {\color[HTML]{9B9B9B} 0.57} & {\color[HTML]{9B9B9B} 0.79} & {\color[HTML]{9B9B9B} 0.27} & {\color[HTML]{9B9B9B} 0.57}    & {\color[HTML]{9B9B9B}0.30}  & {\color[HTML]{9B9B9B}0.09}\\ \midrule
\multicolumn{1}{c|}{\multirow{4}{*}{20\% (10.2 GB)}} & \multicolumn{1}{c|}{SVD}      & 20061                     & \multicolumn{1}{c|}{18800}          & 0.05  & 0.04  & 0.01  & 0.03  & 0.02  & 0.03  & 0.03  & 0.00   & 0.00 \\
\multicolumn{1}{c|}{}                     & \multicolumn{1}{c|}{FWSVD}     & 1727                     & \multicolumn{1}{c|}{1511}                      & 0.09  & 0.11  & 0.05  & 0.08  & 0.10  & 0.05  & 0.08  & 0.00   & 0.00 \\
\multicolumn{1}{c|}{}                     & \multicolumn{1}{c|}{ASVD}      & 11.14                   & \multicolumn{1}{c|}{15.93}                      & 0.29  & 0.53  & 0.64  & 0.41  & 0.68  & 0.17  & 0.45  & 0.21   & 0.04\\ \cmidrule{2-13} 
\multicolumn{1}{c|}{}                     & \multicolumn{1}{c|}{\xin{\sysname \texttt{(W)}}}  & 7.94 (\textcolor{mygreen}{$\downarrow$29\%})  & \multicolumn{1}{c|}{15.84 (\textcolor{mygreen}{$\downarrow$1\%})}       & 0.31  & 0.62  & 0.61 & 0.45  & 0.71  & 0.21  & 0.49 (\textcolor{mygreen}{$\uparrow$9\%})  & 0.26 (\textcolor{mygreen}{+0.05})  & 0.05 (\textcolor{mygreen}{+0.01})\\ 
\multicolumn{1}{c|}{}                     & \multicolumn{1}{c|}{\sysname}  & \textbf{7.73} (\textcolor{mygreen}{$\downarrow$31\%})  & \multicolumn{1}{c|}{\textbf{12.23} (\textcolor{mygreen}{$\downarrow$23\%})}       & \textbf{0.33}  & \textbf{0.67}  & \textbf{0.69 } & \textbf{0.55}  & \textbf{0.79}  & \textbf{0.26}  & \textbf{0.55} (\textcolor{mygreen}{$\uparrow$22\%})  & \textbf{0.28} (\textcolor{mygreen}{+0.07})  & \textbf{0.08} (\textcolor{mygreen}{+0.04})\\ \midrule

\multicolumn{1}{c|}{\multirow{4}{*}{40\% (7.76 GB)}} & \multicolumn{1}{c|}{SVD}      & 52489                    & \multicolumn{1}{c|}{47774}           & 0.04  & 0.04  & 0.05  & 0.01  & 0.03  & 0.02  & 0.03  & 0.00   & 0.00\\
\multicolumn{1}{c|}{}                     & \multicolumn{1}{c|}{FWSVD}     & 18156                     & \multicolumn{1}{c|}{12847}                    & 0.06  & 0.05  & 0.02  & 0.00  & 0.05  & 0.03  & 0.04  & 0.00   & 0.00\\
\multicolumn{1}{c|}{}                     & \multicolumn{1}{c|}{ASVD}      & 1407                 & \multicolumn{1}{c|}{1109}                          & 0.08  & 0.11  & 0.09  & 0.08  & 0.13  & 0.08  & 0.10  & 0.01   & 0.00\\ \cmidrule{2-13} 
\multicolumn{1}{c|}{}                     & \multicolumn{1}{c|}{\xin{\sysname \texttt{(W)}}}  & 13.73 (\textcolor{mygreen}{$\downarrow$99\%}) & \multicolumn{1}{c|}{75.42 (\textcolor{mygreen}{$\downarrow$93\%})}        & 0.25  & 0.33  & 0.55  & 0.40  & 0.63  & 0.12  & 0.38 (\textcolor{mygreen}{$\uparrow$280\%})  & 0.17 (\textcolor{mygreen}{+0.17})   & 0.02 (\textcolor{mygreen}{+0.02})\\ 
\multicolumn{1}{c|}{}                     & \multicolumn{1}{c|}{\sysname}  & \textbf{9.27} (\textcolor{mygreen}{$\downarrow$99\%}) & \multicolumn{1}{c|}{\textbf{15.63} (\textcolor{mygreen}{$\downarrow$99\%})}        & \textbf{0.29}  & \textbf{0.59}  & \textbf{0.68}  & \textbf{0.52}  & \textbf{0.69}  & \textbf{0.20}  & \textbf{0.50} (\textcolor{mygreen}{$\uparrow$400\%})  & \textbf{0.24} (\textcolor{mygreen}{+0.23})   & \textbf{0.07} (\textcolor{mygreen}{+0.07})\\ \midrule
\multicolumn{1}{c|}{\multirow{4}{*}{60\% (5.35 GB)}} & \multicolumn{1}{c|}{SVD}      & 105474                     & \multicolumn{1}{c|}{106976}        & 0.01  & 0.03  & 0.01  & 0.00  & 0.01  & 0.02  & 0.01  & 0.00   & 0.00\\
\multicolumn{1}{c|}{}                     & \multicolumn{1}{c|}{FWSVD}     & 32194                     & \multicolumn{1}{c|}{29292}                    & 0.06  & 0.02  & 0.01  & 0.01  & 0.02  & 0.03  & 0.03  & 0.00   & 0.00\\
\multicolumn{1}{c|}{}                     & \multicolumn{1}{c|}{ASVD}      & 57057                  & \multicolumn{1}{c|}{43036}                       & 0.05  & 0.04  & 0.06  & 0.09  & 0.08  & 0.05  & 0.06  & 0.00   & 0.00\\ \cmidrule{2-13} 
\multicolumn{1}{c|}{}                     & \multicolumn{1}{c|}{\xin{\sysname \texttt{(W)}}}  & 66.62 (\textcolor{mygreen}{$\downarrow$99\%}) & \multicolumn{1}{c|}{471.83 (\textcolor{mygreen}{$\downarrow$99\%})}       & 0.10  & 0.05  & 0.17  & 0.10  & 0.21  & 0.04  & 0.11 (\textcolor{mygreen}{$\uparrow$83\%})  & 0.01 (\textcolor{mygreen}{+0.01})   & 0.00 (\textcolor{mygreen}{+0.00})\\ 
\multicolumn{1}{c|}{}                     & \multicolumn{1}{c|}{\sysname}  & \textbf{15.00} (\textcolor{mygreen}{$\downarrow$99\%}) & \multicolumn{1}{c|}{\textbf{26.26} (\textcolor{mygreen}{$\downarrow$99\%})}       & \textbf{0.18}  & \textbf{0.42}  & \textbf{0.44}  & \textbf{0.31}  & \textbf{0.35}  & \textbf{0.12}  & \textbf{0.30} (\textcolor{mygreen}{$\uparrow$400\%})  & \textbf{0.14} (\textcolor{mygreen}{+0.14})   & \textbf{0.04} (\textcolor{mygreen}{+0.04})\\ \midrule
\multicolumn{1}{c|}{\multirow{4}{*}{80\% (2.58 GB)}} & \multicolumn{1}{c|}{SVD}      & 687291                     & \multicolumn{1}{c|}{708243}        & 0.00  & 0.01  & 0.02  & 0.01  & 0.01  & 0.00  & 0.01  & 0.00   & 0.00\\
\multicolumn{1}{c|}{}                     & \multicolumn{1}{c|}{FWSVD}     & 96872                    & \multicolumn{1}{c|}{89243}                     & 0.01  & 0.02  & 0.00  & 0.01  & 0.01  & 0.00  & 0.01  & 0.00   & 0.00\\
\multicolumn{1}{c|}{}                     & \multicolumn{1}{c|}{ASVD}      & 80425                 & \multicolumn{1}{c|}{67927}                        & 0.04  & 0.03  & 0.03  & 0.02  & 0.01  & 0.01  & 0.03  & 0.00   & 0.00\\ \cmidrule{2-13} 
\multicolumn{1}{c|}{}                     & \multicolumn{1}{c|}{\xin{\sysname \texttt{(W)}}}  & 1349 (\textcolor{mygreen}{$\downarrow$98\%}) & \multicolumn{1}{c|}{6224 (\textcolor{mygreen}{$\downarrow$91\%})}       & 0.07  & 0.03  & 0.04  & 0.02  & 0.07  & 0.01  & 0.04 (\textcolor{mygreen}{$\uparrow$33\%})  & 0.00 (\textcolor{mygreen}{+0.00})  & 0.00 (\textcolor{mygreen}{+0.00})\\ 
\multicolumn{1}{c|}{}                     & \multicolumn{1}{c|}{\sysname}  & \textbf{31.79} (\textcolor{mygreen}{$\downarrow$99\%}) & \multicolumn{1}{c|}{\textbf{43.71} (\textcolor{mygreen}{$\downarrow$99\%})}       & \textbf{0.11}  & \textbf{0.23}  & \textbf{0.21}  & \textbf{0.14}  & \textbf{0.17}  & \textbf{0.08}  & \textbf{0.16} (\textcolor{mygreen}{$\uparrow$433\%})  & \textbf{0.04} (\textcolor{mygreen}{+0.04})  & \textbf{0.02} (\textcolor{mygreen}{+0.02})\\ \midrule
\end{tabular}
}
\vspace{-4mm}
\label{tab:dataset_acc}
\end{table*}
}}

\newcommand{\differntllmaccuracyTable}{{
\begin{table}[t]
\centering
\captionof{table}{Perplexity ($\downarrow$) of \sysname, \sysname \texttt{(W)}, and baselines on WikiText-2 and the average accuracy ($\uparrow$) of the six common sense reasoning datasets of four different LLMs -- OPT-6.7B, LLaMA 2-7B, Mistral-7B, and Vicuna-7B -- under 20\% compression ratio. The relative performance gain compared to the best-performing baseline is marked in green color inside bracket.}
\vspace{-2mm}
\resizebox{1\textwidth}{!}{%
\begin{tabular}{c|cc|cc|cc|cc}
\midrule
& \multicolumn{2}{c}{OPT-6.7B}   & \multicolumn{2}{c}{\textsc{LLaMA 2-7B}}    & \multicolumn{2}{c}{\textsc{Mistral-7B}}        & \multicolumn{2}{c}{\textsc{Vicuna-7B}} \\\midrule
\textsc{Method} & Perplexity\textcolor{mygreen}{$\downarrow$}    & Accuracy\textcolor{mygreen}{$\uparrow$}    & Perplexity\textcolor{mygreen}{$\downarrow$}    & Accuracy\textcolor{mygreen}{$\uparrow$}  & Perplexity\textcolor{mygreen}{$\downarrow$}    & Accuracy\textcolor{mygreen}{$\uparrow$}  & Perplexity\textcolor{mygreen}{$\downarrow$}    & Accuracy\textcolor{mygreen}{$\uparrow$} \\ \midrule
\color[HTML]{9B9B9B}Original       & \color[HTML]{9B9B9B}10.86          & \color[HTML]{9B9B9B}0.52         & \color[HTML]{9B9B9B}5.47            & \color[HTML]{9B9B9B}0.57      & \color[HTML]{9B9B9B}5.25          & \color[HTML]{9B9B9B}0.61         & \color[HTML]{9B9B9B}6.78            & \color[HTML]{9B9B9B}0.56    \\\midrule
SVD       & 66275      & 0.03         & 18192             & 0.09      & 159627        & 0.03         & 18644            & 0.05    \\
FWSVD     & 14559      & 0.06         & 2360              & 0.12      & 6357         & 0.08          & 2758              & 0.09 \\
ASVD      & 82.00      & 0.32         & 10.10             & 0.36      & 13.72         & 0.32         & 16.23             & 0.33     \\ \midrule
\xin{\sysname \texttt{(W)}}  & 16.04 (\textcolor{mygreen}{$\downarrow$80\%})         & 0.41 (\textcolor{mygreen}{$\uparrow$28\%})           &  8.50 (\textcolor{mygreen}{$\downarrow$16\%})          & 0.53 (\textcolor{mygreen}{$\uparrow$47\%})      & 10.21 (\textcolor{mygreen}{$\downarrow$26\%})         & 0.42 (\textcolor{mygreen}{$\uparrow$24\%})           &  8.41 (\textcolor{mygreen}{$\downarrow$48\%})          & 0.51 (\textcolor{mygreen}{$\uparrow$55\%})   \\
\sysname  & \textbf{14.47} (\textcolor{mygreen}{$\downarrow$82\%})         & \textbf{0.49} (\textcolor{mygreen}{$\uparrow$53\%})           &  \textbf{7.73} (\textcolor{mygreen}{$\downarrow$23\%})          & \textbf{0.54} (\textcolor{mygreen}{$\uparrow$50\%})      & \textbf{7.47} (\textcolor{mygreen}{$\downarrow$45\%})         & \textbf{0.55} (\textcolor{mygreen}{$\uparrow$72\%})           &  \textbf{7.43} (\textcolor{mygreen}{$\downarrow$54\%})          & \textbf{0.54} (\textcolor{mygreen}{$\uparrow$64\%})   \\ \midrule
\end{tabular}
}
\vspace{-4mm}
\label{tab:different_llm_acc}
\end{table}
}}

\newcommand{\largellmaccuracyTable}{{
\centering
\captionof{table}{Perplexity ($\downarrow$) of \sysname, \sysname \texttt{(W)}, and baselines on WikiText-2 and the average accuracy ($\uparrow$) of the six classification datasets of LLaMA-13B and LLaMA-30B under 20\% compression ratio. The relative performance gain compared to the best-performing baseline is marked in green color inside bracket.}
\vspace{-2mm}
\resizebox{0.48\textwidth}{!}{%
\begin{tabular}{c|cc|cc}
\midrule
& \multicolumn{2}{c}{\textsc{LLaMA-13B}}    & \multicolumn{2}{c}{\textsc{LLaMA-30B}}         \\\midrule
\textsc{Method}     & Perplexity\textcolor{mygreen}{$\downarrow$}    & Accuracy\textcolor{mygreen}{$\uparrow$}       & Perplexity\textcolor{mygreen}{$\downarrow$}    & Accuracy\textcolor{mygreen}{$\uparrow$}    \\ \midrule
\color[HTML]{9B9B9B}Original                 & \color[HTML]{9B9B9B}5.09        & \color[HTML]{9B9B9B}0.59           & \color[HTML]{9B9B9B}4.10         & \color[HTML]{9B9B9B}0.61 \\\midrule
SVD                 & 946.31        & 0.21           & 54.11         & 0.33            \\
FWSVD               & 15.98         & 0.43           & 20.54         & 0.42             \\
ASVD                & 6.74          & 0.54           & 22.71         & 0.44             \\ \midrule
\xin{\sysname \texttt{(W)}}            &  6.61 (\textcolor{mygreen}{$\downarrow$2\%})         & 0.54 (\textcolor{mygreen}{$\uparrow$0\%})       & 5.63 (\textcolor{mygreen}{$\downarrow$73\%})             & 0.57 (\textcolor{mygreen}{$\uparrow$30\%})          \\
\sysname            &  \textbf{6.43} (\textcolor{mygreen}{$\downarrow$5\%})         & \textbf{0.55} (\textcolor{mygreen}{$\uparrow$2\%})       &\textbf{5.14} (\textcolor{mygreen}{$\downarrow$75\%})             & \textbf{0.59} (\textcolor{mygreen}{$\uparrow$34\%})          \\ \midrule
\end{tabular}
}
\vspace{-5mm}
\label{tab:large_llm_acc}
}}

\newcommand{\loraTable}{{
\centering
\captionof{table}{Perplexity of LLaMA-7B compressed by \sysname under 20\% to 80\% compression ratio on WikiText-2 with different updating orders.}
\vspace{1mm}
\resizebox{1\textwidth}{!}{%
\begin{tabular}{c|cccc}
\midrule
\textsc{Updating Order}     & 20\%      & 40\%      & 60\%   & 80\%   \\ \midrule
$W'_u$ first, then $W'_v$         & 7.85     & 9.32        & \textbf{13.20} (\textcolor{mygreen}{$\downarrow$1\%})  & \textbf{31.67}(\textcolor{mygreen}{$\downarrow$1\%}) \\ \midrule
$W'_v$ first, then $W'_u$          & \textbf{7.73} (\textcolor{mygreen}{$\downarrow$2\%})      & \textbf{9.27} (\textcolor{mygreen}{$\downarrow$1\%})         & 15.00    & 31.79  \\ \midrule
\end{tabular}
}
\label{tab:lora_acc}
}}

\newcommand{\sensitivityTable}{{
\centering
\captionof{table}{Perplexity ($\downarrow$) of compressed LLaMA-7B on WikiText-2 under different compression ratios. \sysname \texttt{(W)} denotes the version of \sysname with truncation-aware data whitening only; \sysname \texttt{(U)} denotes the version of \sysname with parameter update with sequential low-rank approximation only; The relative performance gain compared to ASVD is marked in green color inside bracket.}
\vspace{5mm}
\resizebox{1\textwidth}{!}{%
\begin{tabular}{c|c|c|c}
\midrule
\textsc{Method} & \multicolumn{1}{c}{\textsc{20\%}}  &\multicolumn{1}{c}{\textsc{40\%}}  &\multicolumn{1}{c}{\textsc{60\%}}     \\ \midrule          
ASVD              & 11.14                 & 1407              & 57057                        \\ \midrule
\sysname \texttt{(W)}      & 7.94 (\textcolor{mygreen}{$\downarrow$29\%})            & 13.11 (\textcolor{mygreen}{$\downarrow$99\%})          & 42.30 (\textcolor{mygreen}{$\downarrow$99\%})     \\ \midrule
\sysname \texttt{(U)}      & 10.12 (\textcolor{mygreen}{$\downarrow$9\%})            & 19.28 (\textcolor{mygreen}{$\downarrow$99\%})          & 49.88 (\textcolor{mygreen}{$\downarrow$99\%})     \\ \midrule
\sysname  & \textbf{7.73} (\textcolor{mygreen}{$\downarrow$31\%})      & \textbf{9.27} (\textcolor{mygreen}{$\downarrow$99\%})         & \textbf{15.00} (\textcolor{mygreen}{$\downarrow$99\%})           \\ \midrule
\end{tabular}
}
\vspace{-4mm}
\label{tab:sensitivity}
}}

\newcommand{\svdquantTable}{{
\centering
\captionof{table}{Perplexity ($\downarrow$) of LLaMA-7B compressed by 1-bit quantization methods and \sysname on WikiText-2. The relative performance gain compared to the best-performing baseline is marked in green.}
\resizebox{1\textwidth}{!}{%
\begin{tabular}{c|c|c|c}
\midrule
\textsc{Method}         & \textsc{Type}         &\textsc{Memory} & \textsc{Perplexity}\\ \midrule
PB-LLM                  & Post-training         & 1.9 GB            & 104.83 \\
BiLLM                   & Post-training         & 1.5 GB            & 47.67           \\\midrule
\sysname                & Post-training         & 1.5 GB            & \textbf{47.21} (\textcolor{mygreen}{$\downarrow$1\%})        \\\midrule
OneBit                  & Training-required     & 1.3 GB            & 10.20           \\\midrule
\sysname (2-bit)       & Post-training         & 1.3 GB            & \textbf{9.83} (\textcolor{mygreen}{$\downarrow$4\%})          \\\midrule
\end{tabular}
}
\label{tab:svd_quant}
}}

\newcommand{\svdpruneTable}{{
\centering
\captionof{table}{Perplexity ($\downarrow$) of LLaMA-7B compressed by structured pruning methods and \sysname under various memory budget on WikiText-2. The relative performance gain compared to the best-performing baseline is marked in green.}
\resizebox{1\textwidth}{!}{%
\begin{tabular}{c|c|c|c|c}
\midrule
& \multicolumn{4}{c}{\textsc{Perplexity under various memory budget}}\\ \midrule
\textsc{Method}       & 10 GB       & 9 GB      & 8 GB    & 7 GB \\ \midrule
LLM-Pruner            & 9.88        & 12.21      & 18.94     & 21.68 \\
SliceGPT              & 8.78        & 12.73      & 16.39     & 27.41           \\
BlockPruner           & 9.40         & 12.76      & 19.78     & 43.05        \\\midrule
\sysname              & \textbf{7.92} (\textcolor{mygreen}{$\downarrow$10\%})        & \textbf{8.18} (\textcolor{mygreen}{$\downarrow$33\%})      & \textbf{8.33} (\textcolor{mygreen}{$\downarrow$49\%})     & \textbf{9.63} (\textcolor{mygreen}{$\downarrow$56\%})        \\\midrule
\end{tabular}
}
\label{tab:svd_prune}
}}

\newcommand{\timeTable}{{
\begin{table}[ht]
\centering
\captionof{table}{Compression time of \sysname and ASVD on LLaMA-7B under 20\% compression ratio. The relative speedup is marked in green color inside bracket.}
\vspace{-2mm}
\resizebox{0.9\textwidth}{!}{%
\begin{tabular}{ccc|ccc}
\midrule
 \multicolumn{3}{c|}{\sysname} & \multicolumn{3}{c}{ASVD}       \\ \cmidrule{1-6} 
                         \makecell{Truncation-Aware \\Data Whitening}  & \makecell{Parameter Update with Sequential \\ Low-rank Approximation}  & \textbf{Total} & Normalize & Search & \textbf{Total} \\ \midrule
                   10min      & 3.5h    & \textbf{3.5h} (\textcolor{mygreen}{$\downarrow$36\%}) & 5min          & 5.5h   & \textbf{5.5h}  \\ \midrule
\end{tabular}
}
\label{tab:time}
\end{table}
}}
\newcommand{\datasetTable}{{
\centering
\captionof{table}{Performance of LLaMA-7B compressed by \sysname under 20\% compression ratio using calibration data sampled from WikiText-2 (by default in our paper) and C4 datasets. The performance on WikiText-2 and C4 are reported by perplexity ($\downarrow$), while the performance on six downstream datasets are reported by average accuracy ($\uparrow$). The performance on TruthfulQA and GSM8K are reported by BLEU score($\uparrow$) and Exact Match Accuracy ($\uparrow$) respectively. 
The relative performance gain for data sampled from one dataset compared to another is marked in green color inside bracket.
}
\vspace{-2.5mm}
\resizebox{1\textwidth}{!}{%
\begin{tabular}{cc|c|cc}
\midrule
WikiText-2\textcolor{mygreen}{$\downarrow$} & C4\textcolor{mygreen}{$\downarrow$} & Average\textcolor{mygreen}{$\uparrow$}  & TruthfulQA\textcolor{mygreen}{$\uparrow$}  & GSM8K\textcolor{mygreen}{$\uparrow$}\\ \midrule
\multicolumn{5}{c}{Calibration data sampled from WikiText-2 } \\ \midrule
 \textbf{7.73} (\textcolor{mygreen}{$\downarrow$1\%})       & 12.23      & \textbf{0.55} (\textcolor{mygreen}{$\uparrow$2\%})     & 0.28   & 0.08               \\\midrule
\multicolumn{5}{c}{Calibration data sampled from C4} \\ \midrule
     7.79      & \textbf{11.97} (\textcolor{mygreen}{$\downarrow$1\%})      & 0.54     & 0.28   & 0.08               \\\midrule
\end{tabular}
}
\label{tab:different_ds_acc}
}}

\newcommand{\scalingTable}{{
\begin{table*}[t]
\centering
\vspace{6mm}
\captionof{table}{Comparison of LLaMA-3B (compressed from LLaMA-7B by \sysname) and original StableLM-3B~\citep{StableLM-3B-4E1T} trained from scratch. Both the throughput and the peak memory footprint during the inference are measured with batch size=32, sequence length = 128 on single A100 GPU. }
\resizebox{1\textwidth}{!}{
\begin{tabular}{c|c|c|cccccc|c|c|c}
\midrule
\textsc{Model}      & Throughput        & Peak Mem. & Openb.  & Arc\_e  & WinoG. & HellaS.   & PIQA      & MathQA    & \textbf{Average\textcolor{mygreen}{$\uparrow$}}  & TruthfulQA\textcolor{mygreen}{$\uparrow$} & GSM8K\textcolor{mygreen}{$\uparrow$} \\ \midrule
StableLM-3B         & 8463 Tokens/sec             &  9.41 GB     & 0.19   & 0.51   & 0.55    & 0.47      & 0.69      & 0.21      & 0.44  & 0.22  & 0.02                  \\ \midrule
LLaMA-3B            & 9254 Tokens/sec             &  7.43 GB     & \textbf{0.27}      & \textbf{0.54}    & \textbf{0.60}    & \textbf{0.49}      & 0.68      & 0.19      & \textbf{0.46} (\textcolor{mygreen}{$\uparrow$5\%})   & \textbf{0.23} (\textcolor{mygreen}{+ 0.01})   & \textbf{0.04} (\textcolor{mygreen}{+ 0.02})                  \\ \midrule
\end{tabular}
}
\label{tab:scaling}
\end{table*}
}}

\newcommand{\exampleLoraTable}{{
\begin{table}[ht]
    \centering
    \vspace{-3mm}
    \caption{An example of contents generated by the fine-tuned LLaMA-7B with LoRA after being compressed by \sysname and ASVD under different compression ratios. The input is marked in bold and the normal texts are the generated sentences.} \label{tab:visualization_lora}
    \resizebox{\linewidth}{!}{
    \tiny
    \begin{tabular}{p{0.8cm}|p{5cm}|p{4.8cm}}
        \midrule
        \textsc{Ratio} & \textsc{\sysname} & \textsc{ASVD} \\
        \midrule
        Original & \multicolumn{2}{p{10.2cm}}{\textbf{What is the responsibility of the AI assistant?} The responsibility of an AI assistant is to complete the job without any hesitation. Also, the AI assistant has no problem with taking a decision and executing it. An AI assistant is responsible to complete the task and it needs to learn from his past experience to take the best possible decision without any hesitation...} \\
        \midrule
         20\% & \textbf{What is the responsibility of the AI assistant?} The AI assistant's role and responsibilities are often ambiguous and can vary depending on the context and the type of assistant being used. Some AI assistants may have preset responses to common queries, but their behavior can also depend on the user's interaction and expectations...  & \textbf{What is the responsibility of the AI assistant?} 1. To act in the same way as the programmer intended. 2. To make a value assessment of the user's intentions and offer an appropriate response. 3. To help the user carry out specific tasks or follow a strict script. 4. To make the system as independent of the user as possible...\\
        \midrule
        40\% & \textbf{What is the responsibility of the AI assistant?} 1. The AI assistant is responsible for performing tasks that can only be done by AI technology such as scheduling appointments, setting reminders and making suggestions, while also providing quick and accurate responses. 2. The AI assistant is also responsible for learning and updating its knowledge base through continuous improvement, thus making it more efficient and effective. 3. The AI assistant is responsible for handling and managing user-generated and customer-oriented input... 
        & \textbf{What is the responsibility of the AI assistant?}  1. Define the definition of the job specification. 2. Define the role of the AI assistant for the job. 3. Define the skills necessary in order to carryout the tasks assigned to the job to do the work of the same capacity in the market. 4. Determine the evaluation of the job competence of the AI assistant to ensure their professional integrity. 5. What type of AI assistant jobs are available? 6. What industries require technology assistance work to complete the tasks that it own personnel cannot perform...\\
        \midrule
        60\% & \textbf{What is the responsibility of the AI assistant?} In a nutshell, the responsibility of the AI assistant could vary depending on the task, but generally, the focus would be on automatic tasks, without the need for human intervention. Some common tasks could include setting reminders, scheduling appointments, and making routine household or productivity tasks. The AI assistant also serves as a backup or a relief system, taking on responsibilities when the user is not available ... 
        & \textbf{What is the responsibility of the AI assistant?}  2.3. ?? Brush for a discussion I wonder is it worth doing is important.2- It isn't useful just reducing labor costs; it helps employees feel a sense of connected to your attention which leads to better workplace values among staffers and leads to long relationships among org...\\
        \midrule
        80\% & \textbf{What is the responsibility of the AI assistant?}  Our Design is based on our understanding of the world, and we are actively learning, adapting and adapting, so we're always evolving new ideas, which we see to be most unique and relevant in our community... 
        & \textbf{What is the responsibility of the AI assistant?} ygua AIeltemperaturen/2, (64mbz/.3/.1/, 7.kbld.org.0/2/ In these  puthebout les bnvols n merginels ...\\
        \midrule
    \end{tabular}
    }
\end{table}
}}

\newcommand{\flapTable}{{
\begin{table}[t]
\centering
\captionof{table}{Perplexity ($\downarrow$) of \sysname and FLAP on WikiText-2 to compress LLaMA-7B under different compression ratios. The better performance is marked in bold. The relative performance gain of \sysname compared to FLAP is marked in green inside bracket.}
\vspace{-2mm}
\resizebox{0.8\textwidth}{!}{%
\begin{tabular}{c|c|c|c|c}
\midrule
\textsc{Ratio (Mem.)} & \textsc{20\% (10.2GB)}   & \textsc{40\% (7.76GB)}    & \textsc{60\% (5.35GB)}        & \textsc{80\% (2.58GB)} \\\midrule
FLAP      & 7.99              & 14.43             & 106.87      & 15023         \\ \midrule
\sysname  & \textbf{7.73} (\textcolor{mygreen}{$\downarrow$3\%})    &  \textbf{9.27} (\textcolor{mygreen}{$\downarrow$36\%})               & \textbf{15.00} (\textcolor{mygreen}{$\downarrow$86\%})            &  \textbf{31.79} (\textcolor{mygreen}{$\downarrow$99\%})             \\ \midrule
\end{tabular}
}
\label{tab:flap}
\end{table}
}}

\newcommand{\loraaddTable}{{
\begin{table}[t]
\centering
\captionof{table}{Perplexity ($\downarrow$) of \sysname with original LoRA fine-tuning (denoted as \sysname (SFT)), ASVD with sequential LoRA fine-tuning (denoted as ASVD (SFT)), and  \sysname with sequential LoRA fine-tuning (denoted as \sysname (SFT)) on WikiText-2 to compress LLaMA-7B under different compression ratios.}
\vspace{-2mm}
\resizebox{0.8\textwidth}{!}{%
\begin{tabular}{c|c|c|c|c}
\midrule
\textsc{Ratio (Mem.)} & \textsc{20\% (10.2GB)}   & \textsc{40\% (7.76GB)}    & \textsc{60\% (5.35GB)}        & \textsc{80\% (2.58GB)} \\\midrule
\sysname (NFT)  & 7.87            & 11.98             & 16.30      & 80.23         \\ 
ASVD (SFT)  & 8.37            & 14.86             & 44.81      & 271         \\ \midrule
\sysname (SFT)  & \textbf{7.73}    &  \textbf{9.27}               & \textbf{15.00}          &  \textbf{31.79}           \\ \midrule
\end{tabular}
}
\vspace{-5mm}
\label{tab:lora_add}
\end{table}
}}

\newcommand{\dronelossTable}{{
\begin{table}[t]
\centering
\captionof{table}{Compression loss of the randomly generated weight and activation matrices with different shapes under 50\% compression ratio using \sysname, Drone, and the theoretical minimum.}
\vspace{-2mm}
\resizebox{0.9\textwidth}{!}{%
\begin{tabular}{c|c|c|c}
\midrule
\textsc{Loss} & $[128\times 128] \times [128\times128]$ & $[2048\times 2048] \times [2048\times 2048]$ & $[4096\times 4096] \times [4096\times4096]$ \\\midrule
\textsc{Minimum}    &   276.1130          &     17784.2637      &   50321.9141              \\\midrule
\textsc{Drone}      &   276.1130          &     17785.6992     &   50337.2148       \\\midrule
\sysname            &   276.1130          &     17784.2676      &   50321.9727       \\\midrule
\end{tabular}
}
\vspace{-4mm}
\label{tab:drone_loss}
\end{table}
}}

\newcommand{\dronetimeTable}{{
\begin{table}[t]
\centering
\captionof{table}{Compression Time of the randomly generated weight and activation matrices with different shapes using \sysname and Drone. The compression time is measured for 5 times' compression.}
\vspace{-2mm}
\resizebox{0.9\textwidth}{!}{%
\begin{tabular}{c|c|c|c}
\midrule
\textsc{Time} & $[128\times 128] \times [128\times128]$ & $[2048\times 2048] \times [2048\times 2048]$ & $[4096\times 4096] \times [4096\times4096]$
\\\midrule
\textsc{Drone}  &       0.07 seconds          &   5.81 seconds        &       30.35 seconds          \\\midrule
\sysname        &       0.02 seconds          &   1.98 seconds        &       10.37 seconds          \\\midrule
\end{tabular}
}
\vspace{-5mm}
\label{tab:drone_time}
\end{table}
}}
\usepackage{microtype}
\usepackage{graphicx}
\usepackage{subfigure}
\usepackage{caption}
\usepackage{xspace}
\usepackage{pifont}
\usepackage{makecell}
\usepackage{bm}

\usepackage{subcaption}
\usepackage{booktabs} 
\usepackage{wrapfig}

\PassOptionsToPackage{numbers, compress}{natbib}
\usepackage{iclr2025_conference,times}

\usepackage{amsmath,amsfonts,bm}

\def\eqref#1{equation~\ref{#1}}

\def\1{\bm{1}}

\DeclareMathAlphabet{\mathsfit}{\encodingdefault}{\sfdefault}{m}{sl}
\SetMathAlphabet{\mathsfit}{bold}{\encodingdefault}{\sfdefault}{bx}{n}

\usepackage[utf8]{inputenc} 
\usepackage[T1]{fontenc}    
\usepackage{hyperref}       
\usepackage{url}            
\usepackage{amsfonts}       
\usepackage{nicefrac}       
\usepackage{xcolor}         

\usepackage{amsmath}
\usepackage{amssymb}
\usepackage{mathtools}
\usepackage{amsthm}
\usepackage{multirow}
\usepackage{graphicx}
\usepackage{lscape}
\usepackage{algorithm}
\usepackage{algpseudocode}
\usepackage[capitalize,noabbrev]{cleveref}

\theoremstyle{plain}
\newtheorem{theorem}{Theorem}[section]

\newtheorem{lemma}[theorem]{Lemma}
\newtheorem{corollary}[theorem]{Corollary}
\theoremstyle{definition}

\theoremstyle{remark}

\newcommand{\sysname}{\texttt{SVD-LLM}\xspace}

\usepackage[textsize=tiny]{todonotes}

\title{\raisebox{-0.85ex}{\includegraphics[width=1.5em]{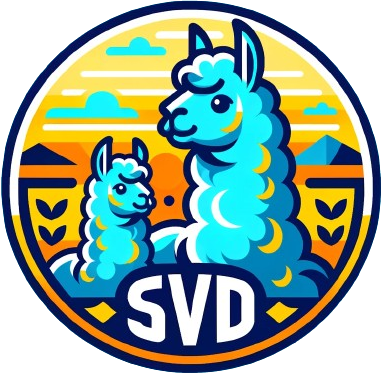}}\hspace{0.3em}SVD-LLM: Truncation-aware Singular Value Decomposition for Large Language Model Compression}

\author{%
 Xin Wang\textsuperscript{\rm 1} \quad
 Yu Zheng\textsuperscript{\rm 2} \quad 
 Zhongwei Wan\textsuperscript{\rm 1} \quad 
 Mi Zhang\textsuperscript{\rm 1} \\
 \textsuperscript{\rm 1}The Ohio State University  \quad
 \textsuperscript{\rm 2}Michigan State University\\
 \url{https://github.com/AIoT-MLSys-Lab/SVD-LLM} \\
}

\usepackage{xcolor}

\newcommand{\xin}[1]{\textcolor{black}{#1}}

\iclrfinalcopy
\begin{document}

\definecolor{mygreen}{HTML}{009901}
\definecolor{myred}{HTML}{A52A2A}

\maketitle
 
\begin{abstract}
The advancements in Large Language Models (LLMs) have been hindered by their substantial sizes, which necessitates LLM compression methods for practical deployment. Singular Value Decomposition (SVD) offers a promising solution for LLM compression. However, state-of-the-art SVD-based LLM compression methods have two key limitations: truncating smaller singular values may lead to higher compression loss, and the lack of update on the compressed weights after SVD truncation. 
In this work, we propose \sysname, a SVD-based  post-training LLM compression method that addresses the limitations of existing methods. \sysname incorporates a truncation-aware data whitening technique to ensure a direct mapping between singular values and compression loss. Moreover, \sysname adopts a parameter update with sequential low-rank approximation to compensate for the accuracy degradation after SVD compression. 
We evaluate \sysname on $10$ datasets and seven models from three different LLM families at three different scales. Our results demonstrate the superiority of \sysname over state-of-the-arts, especially at high model compression ratios.
%

\end{abstract}

\section{Introduction}
\label{sec:introduction}
\vspace{-3mm}
Large Language Models (LLMs) have demonstrated remarkable capabilities in a wide range of tasks such as natural language understanding and generation~\citep{DBLP:journals/corr/abs-2303-18223, DBLP:journals/corr/abs-2306-02781}. 
Despite such capabilities, the democratization of LLMs is primarily restricted by their substantial resource demands, which motivates the design of LLM compression methods~\citep{DBLP:journals/tmlr/Wan0LA0LQYZZC024, DBLP:journals/corr/abs-2401-01923, DBLP:journals/tacl/ZhuLLMW24, DBLP:journals/corr/abs-2404-14294}.
These methods are often performed in a post-training manner without requiring retraining from scratch.
%
Post-training LLM compression methods based on quantization~\citep{DBLP:conf/iclr/YuanSD24, DBLP:conf/icml/HuangLQLZ0M024}, unstructured pruning~\citep{DBLP:conf/icml/FrantarA23}, and structured pruning~\citep{DBLP:conf/nips/MaFW23, DBLP:conf/iclr/AshkboosCNHH24, DBLP:journals/corr/abs-2406-10594} have been intensively studied. 
Despite their success, these methods have certain limitations, such as dependence on specific hardware and low inference speedup. In contrast, compression methods based on low-rank approximation, such as Singular Value Decomposition (SVD) are not limited by those constraints. Moreover, the KV cache of LLMs compressed via SVD at runtime can also be reduced.

Despite these advantages, the potential of SVD for LLM compression has not been fully explored. Several SVD-based LLM compression methods, such as FWSVD~\citep{DBLP:conf/iclr/HsuHCLSJ22} and ASVD~\citep{DBLP:journals/corr/abs-2312-05821} have been proposed. However, these methods exhibit severe performance degradation when model compression ratio\footnote{Model compression ratio refers to the percentage of parameter reduction achieved through compression.} increases. 
Such limitation can be attributed to two fundamental issues involved in their approaches.
\ding{182} \textbf{Misalignment between SVD truncation and compression loss}:
%
both FWSVD and ASVD fail to establish a direct relationship between singular values and model compression loss. As a consequence, truncating smaller singular values in SVD could lead to higher compression loss.
\ding{183} \textbf{Lack of model parameter update after SVD truncation}: 
as model compression ratio increases, the number of singular values that need to be truncated in SVD increases as well. To compensate for the accuracy degradation caused by truncating a larger number of singular values, it becomes necessary to update the remaining parameters of the compressed model.
Unfortunately, existing SVD-based LLM compression methods do not incorporate such update in their design, and thus fail to compensate for the accuracy degradation especially under high model compression ratios. 


In this paper, we propose a SVD-based post-training LLM compression method, \sysname, which effectively addresses the two fundamental issues of existing SVD-based LLM compression methods.
\sysname differs from them in two key aspects.
%
\ding{182} \textbf{Truncation-Aware Data Whitening:}
supported by theoretical proof, \sysname incorporates a truncation-aware data whitening technique that ensures a \textit{direct mapping} between singular values and model compression loss. In doing so, the proposed truncation-aware data whitening technique is able to identify which singular values should be truncated to incur minimal model compression loss. 
%
\ding{183} \textbf{Parameter Update with Sequential Low-rank Approximation:}
to compensate for accuracy degradation after compression, \sysname sequentially fine-tunes the decomposed low-ranking matrices for a global accuracy recovery.

We compare \sysname with both state-of-the-art SVD-based LLM compression methods as well as pruning and quantization-based LLM compression methods. To demonstrate the generability of \sysname, we conduct our evaluation on a total of $10$ datasets and seven models from three different LLM families (LLaMA, OPT, and Mistral) at three different scales (7B, 13B, 30B), and evaluate the performance of \sysname on both GPU and CPU. We highlight three of our findings:

\vspace{-2mm}
\begin{itemize}
    \item \sysname outperforms state-of-the-art SVD-based LLM compression methods FWSVD and ASVD across all $10$ datasets, three LLM families at three scales by a large margin.
    \item  \sysname also outperforms state-of-the-art structured pruning-based LLM compression methods, including LLM-Pruner, SliceGPT, BlockPruner as well as state-of-the-art 1-bit post-training quantization-based LLM compression methods, including PB-LLM and BiLLM. More importantly, when combined with 2-bit post-training quantization, \sysname outperforms state-of-the-art 1-bit training-required quantization-based LLM compression method OneBit, presenting a new way to achieve state-of-the-art compression performance without incurring expensive retraining.
    \item LLMs compressed by \sysname are able to achieve inference speedup and memory reduction when deployed on real hardware, including both GPU and CPU. 
    At the same time, \sysname is able to reduce runtime KV cache memory without additional accuracy drop.
\end{itemize}

\vspace{-1mm}
\section{Related Work}
\label{sec:related_works}
\vspace{-1mm}
\textbf{Large Language Model Compression:}
LLMs in general contain billion-scale parameters. Applying conventional model compression methods for LLMs is unfeasible as they necessitate resource-intensive retraining. Given that, post-training methods that avoid retraining in the compression process have been proposed. 
In general, these methods can be grouped into four categories: unstructured pruning, structured pruning, quantization, and low-rank approximation.
Specifically, unstructured pruning methods~\citep{DBLP:conf/icml/FrantarA23} set the individual weights of an LLM to zero without changing its shape. However, irregular sparsification of unstructured pruning is difficult to achieve the desired speedup or memory saving.
%
Unlike unstructured pruning, structured pruning methods~\citep{DBLP:conf/nips/MaFW23, DBLP:conf/iclr/AshkboosCNHH24, DBLP:journals/corr/abs-2406-10594} remove entire channels or other structured components from LLMs, making them easier to implement on hardware. One notable contribution is LLM-Pruner~\citep{DBLP:conf/nips/MaFW23}, which groups weight matrices based on their dependency and assigns the pruning ratio to each group based on the estimated importance. 
%
Quantization methods~\citep{DBLP:journals/corr/abs-2405-04532} compress models by reducing the precision of weight matrices of the LLM. However, similar to unstructured pruning, quantization is also difficult to achieve the desired inference speedup due to the lack of hardware support and efficient kernels for low-precision computation~\citep{DBLP:journals/corr/abs-2405-04532}.
Recent studies including PB-LLM~\citep{DBLP:conf/iclr/YuanSD24} and BiLLM~\citep{DBLP:conf/icml/HuangLQLZ0M024} push the frontier to 1-bit quantization. Nevertheless, these approaches often lead to severe accuracy degradation.

\begin{figure*}[t]
\centering
\includegraphics[width=1\textwidth]{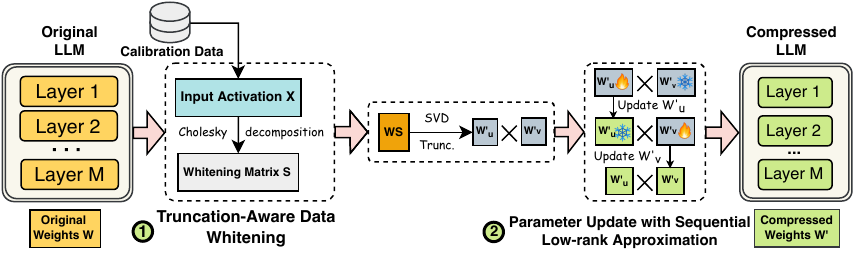}
\vspace{-3mm}
\caption{Overview of \sysname.}
\vspace{-3mm}
\label{fig:framework}
\end{figure*}

\textbf{SVD for Language Model Compression:} Singular Value Decomposition (SVD) is a widely used low-rank approximation technique to reduce matrix size by approximating a matrix with two smaller low-ranking matrices~\citep{GOLUB1987317}. Given that, SVD is commonly used for model compression. 
For instance, DRONE~\citep{DBLP:conf/nips/ChenYDH21}  achieves optimal SVD compression for small language models such as BERT. However, during SVD compression, DRONE caches all the input activations, making it challenging for LLM compression due to excessive memory usage.
%
For LLMs, directly applying SVD on the weight matrix without considering the importance of the weights leads to a large compression loss. To address this issue, \citet{DBLP:conf/iclr/HsuHCLSJ22} propose FWSVD, which introduces Fisher information to weigh the importance of parameters. However, FWSVD requires a complex gradient calculation that demands substantial computing and memory resources for LLM compression. 
Another issue of directly applying SVD is that the distribution of activation can affect the compression loss. To address it, \citet{DBLP:journals/corr/abs-2312-05821} propose ASVD, which scales the weight matrix by a diagonal matrix that normalizes the impact of input channels on the weights.
However, both FWSVD and ASVD do not establish a direct relationship between singular values and compression loss. As a consequence, truncating the smaller singular values may lead to higher compression loss. Moreover, as compression ratio increases, it becomes necessary to update the compressed weights for accuracy recovery. However, both FWSVD and ASVD do not take it into consideration, and thus incur severe accuracy degradation under high compression ratios.

\vspace{-1mm}
\section{SVD-LLM}
\label{sec:methodology}
\vspace{-1mm}
%
\cref{fig:framework} provides an overview of \sysname. 
At a high level, \sysname is a SVD-based post-training LLM compression method.
%
Specifically, following the standard procedure of post-training LLM compression methods~\citep{DBLP:conf/icml/FrantarA23,DBLP:journals/corr/abs-2312-05821,DBLP:conf/icml/XiaoLSWDH23},
\sysname uses a random set of sentences as calibration data to generate activation for truncation-aware data whitening. 
%
Given the generated activation, \sysname derives the whitening matrix $S$ through Cholesky decomposition, and then performs SVD to truncate the multiplication of weight matrices $W$ and whitening matrix $S$ to compress the LLM.
After truncation, \sysname updates the remaining model parameters with sequential low-rank approximation to recover accuracy.
In the following, we describe both truncation-aware data whitening and parameter update with sequential low-rank approximation in detail. The pseudocode of \sysname is provided in~\cref{appendix:pseudocode}.

\subsection{Truncation-Aware Data Whitening}
\label{subsec:activation_aware_weight_whitening}
\textbf{Motivation:}
Due to high variance of input activation, simply applying vanilla SVD for LLM compression leads to severe accuracy degradation~\citep{DBLP:journals/corr/abs-2312-05821}. To address this issue, existing methods~\citep{DBLP:journals/corr/abs-2312-05821, DBLP:conf/iclr/HsuHCLSJ22} formulate LLM compression as an optimization problem with the following objective:
\vspace{-0mm}
\begin{equation}
    O = \min(||WX-W'X||_F) 
    \label{eq:t1}
\end{equation}
       
where $W$ is the weight matrix of the original LLM, $X$ is the activation of $W$, $W'$ is the compressed weight matrix, and $||WX-W'X||_F$ is the compression loss in the form of Frobenius loss.

Although existing methods attempt to reduce this compression loss during their SVD truncation, they all fail to establish a direct relationship between singular values and compression loss. As a consequence, truncating smaller singular values in SVD could lead to significant compression loss. 
Taking ASVD~\citep{DBLP:journals/corr/abs-2312-05821} as an example, ASVD extracts a diagonal matrix $S_{0}$ from $X$ where each element in the diagonal is the absolute mean value of each channel.
It then uses $S_{0}$ to normalize $X$ and converts $WX$ into $(WS_{0})(S_{0}^{-1}X)$. 
Subsequently, SVD is performed on $WS_{0}$ to obtain the decomposed matrices $U_{0}$, $\Sigma_{0}$, and $V_{0}$. Lastly, ASVD truncates the smallest singular values in $\Sigma_{0}$ to obtain the compressed weight matrix $W'_{0} = U_{0} \times \text{Trunc.}(\Sigma_{0}) \times V_{0} \times S_{0}^{-1}$.

\begin{figure}[t]

\centering
\subfigure[Data Normalization (ASVD)]{
\includegraphics[width=0.45\textwidth]{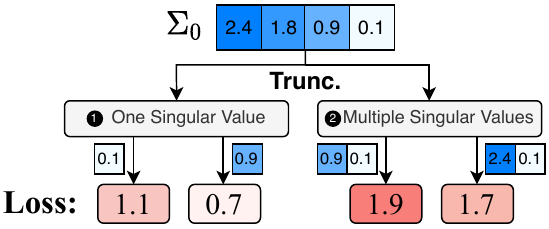}
\label{fig:asvd_whitening}
}
\vline
\hspace{0.1cm}
\centering
\subfigure[Truncation-Aware Data Whitening (\sysname)]{
\includegraphics[width=0.45\textwidth]{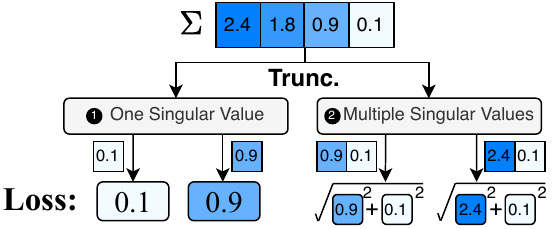}
\label{fig:svdllm_whitening}
}
\DeclareGraphicsExtensions.
\caption{Compression loss ($L = ||WX-W'X||_F$) of different data preprocessing methods. 
}
\vspace{-2mm}
\label{fig:whitening_comparison}
\end{figure}

Although normalizing the activation improves performance, ASVD does not establish a direct relationship between singular values and compression loss (a detailed proof is included in~\cref{subsec:ASVD_reconstruction_error}).
To better illustrate this point, we show two concrete examples in~\cref{fig:asvd_whitening}. 
In example \ding{182} where only one singular value is truncated, truncating the smallest singular value 0.1 results in a higher compression loss (loss $=$ 1.1) compared to truncating the second smallest singular value 0.9 (loss $=$ 0.7). 
In example \ding{183} where multiple singular values are truncated, truncating the smallest two singular values 0.9 and 0.1 also leads to a higher loss (loss $=$ 1.9) than truncating 2.4 and 0.1 (loss $=$ 1.7).
As such, truncating the smallest singular values does not lead to minimal loss.


\textbf{Key Design:}
The key idea of \sysname is to incorporate a truncation-aware data whitening technique that ensures a \textit{direct} mapping between singular values and compression loss. To achieve this, \sysname enforces the whitened activation $S^{-1}X$ to be orthonormal such that each channel is independent of each other, i.e., $(S^{-1}X)(S^{-1}X)^T= S^{-1}XX^T(S^{-1})^T =I$, where $S$ is derived through Cholesky decomposition~\citep{DBLP:books/siam/Meyer00}. 
SVD is then performed on $WS$ to obtain the decomposed matrices $U, \Sigma, V$, where $U=[u_1,u_2,u_3,...,u_r]$, $\Sigma=\text{diag}(\sigma_1,\sigma_2,\sigma_3, \cdots, \sigma_r)$, and $V=[v_1,v_2,v_3,...,v_r]$. Lastly, the smallest singular values in $\Sigma$ are truncated (denoted by $\boldsymbol{\operatorname{Trunc.}}(\Sigma)$) to obtain two low-ranking matrices $W'_u = U\times [\boldsymbol{\operatorname{Trunc.}}(\Sigma)] ^\frac{1}{2}, W'_v = [\boldsymbol{\operatorname{Trunc.}}(\Sigma)] ^\frac{1}{2}\times  V^T \times S^{-1}$ and the compressed weight matrix $W' = W'_u \times W'_v = U \times \boldsymbol{\operatorname{Trunc.}}(\Sigma) \times V^T \times S^{-1}$. 
%

\cref{fig:svdllm_whitening} illustrates the effect of the proposed truncation-aware data whitening method. In example \ding{182} where only one singular value is truncated, the compression loss equals to the truncated singular value. In example \ding{183}, the compression loss of truncating multiple singular values equals to the square root of the sum of their squares. As such, under the proposed truncation-aware data whitening technique, truncating the smallest singular values leads to minimal compression loss.

In the following, we provide a theoretical proof on why the proposed  truncation-aware data whitening technique ensures a direct mapping between singular values and compression loss in the case of one singular value (\cref{thm:whitening}) and multiple singular values (\cref{col:whitening}), respectively.

\vspace{0mm}
\begin{lemma}
\label{lem:frobenius}
    The Frobenius norm of matrix $A$ with dimension $m\times n$ can be deduced into the square root of the trace of its gram matrix, which is: 
        \begin{equation}
        \|A\|_F \triangleq\left(\sum_{j=1}^n \sum_{i=1}^m\left|a_{i j}\right|^2\right)^{\frac{1}{2}} =\left[\boldsymbol{\operatorname{Trace}}\left(A^T A\right)\right]^{\frac{1}{2}}
        \end{equation}
\end{lemma}
Let $\boldsymbol{\operatorname{SVD}}(WS)$ denote SVD compression on matrix $WS$. The compressed weight matrix $W'$ can be expressed as $W' = \boldsymbol{\operatorname{SVD}}(WS)S^{-1}$. Using \cref{lem:frobenius}, we obtain the compression loss $L_i$ when truncating the $i^{th}$ singular value of $WS$ to reduce its rank for compression:
\begin{equation}
\label{formula:loss}
    \begin{split}
    L_i&=\left\|W X-W^{\prime} X\right\|_F=\left\|W S S^{-1} X-\boldsymbol{\operatorname{SVD}}(W S) S^{-1} X\right\|_F=\left\|(W S-\boldsymbol{\operatorname{SVD}}(W S)) S^{-1} X\right\|_F\\
    &=\left\|\sigma_i u_i v_i^T S^{-1} X\right\|_F=\sigma_i \boldsymbol{\operatorname{Trace}}\left(u_i v_i^T S^{-1} X X^T\left(S^{-1}\right)^T v_i u_i^T\right)^{\frac{1}{2}}
    \end{split}
\end{equation}
Since both $U=[u_1,u_2,u_3,...,u_r]$ and $V=[v_1,v_2,v_3,...,v_r]$ are orthonormal matrices, we have:
\begin{equation}
v_i^T v_i = u_i^T u_i = 1; v_i^T v_j = u_i^T u_j = 0, \forall i \neq j ; \boldsymbol{\operatorname{Trace}}(v_i v_i^T) = \boldsymbol{\operatorname{Trace}}(u_i u_i^T) =1
\end{equation}

\begin{theorem}
\label{thm:whitening}
If $S$ is the Cholesky decomposition of $XX^T$, the compression loss $L_i$ equals to $\sigma_i$.
\end{theorem}
\begin{proof}
\label{proof:whitening}
Since the whitening matrix $S$ is the Cholesky decomposition of $XX^T$, we have $SS^T=XX^T$. We can further infer \cref{formula:loss} to obtain:
\begin{equation}
L_i = \sigma_i \boldsymbol{\operatorname{Trace}}(u_i v_i^T v_i u_i^T)^{\frac{1}{2}} = \sigma_i \boldsymbol{\operatorname{Trace}}\left(u_i\left(v_i^T v_i\right) u_i^T\right)^{\frac{1}{2}}=\sigma_i \boldsymbol{\operatorname{Trace}}\left(u_i u_i^T\right)^{\frac{1}{2}} = \sigma_i \\
\end{equation}
Therefore, $L_i$ of truncating $\sigma_i$ equals to the singular value $\sigma_i$ itself.
\end{proof}
\begin{corollary}
\label{col:whitening}
If $S$ is the Cholesky decomposition of $XX^T$, truncating the smallest singular values leads to the lowest loss $L$ compared to truncating others.
\end{corollary}

\begin{proof}
\label{proof:truncation}
If we truncate $\sigma_{m+1}, \sigma_{m+2}, \sigma_{m+3},..., \sigma_{r}$ in $\Sigma$ for compression, the square of the loss $L$ is:
\begin{equation}
    \begin{split}
    L^2 =& \left|\left| \sum_{i=m+1}^r \sigma_i u_i v_i^T S^{-1}X \right|\right|^2_F =\sum_{j=m+1}^r\sum_{i=m+1}^r\sigma_i\sigma_j\boldsymbol{\operatorname{Trace}}(u_i v_i^T S^{-1}XX^T(S^{-1})^T v_j u_j^T) \\
=& \sum_{i=m+1}^r\sigma_i^2\boldsymbol{\operatorname{Trace}}(u_i v_i^T S^{-1}XX^T(S^{-1})^T v_i u_i^T) = \sum_{i=m+1}^r(L_i )^2=\sum_{i=m+1}^r(\sigma_i)^2
    \end{split}
\end{equation}
The squared loss $L^2$ equals to the sum of the squared singular values (More detailed derivation is in Appendix~\ref{appendix:compression_loss_svdllm}). Truncating the smallest singular values achieves the lowest compression loss.
\end{proof}

Given that our proposed truncation-aware data whitening technique is built upon SVD, whose applicability depends on certain singular value distribution, we further conduct a spectrum analysis of the singular values obtained by our technique. We refer the readers to Appendix~\ref{appendix:spectrum_analysis} for details.

It should also be noted that our proposed truncation-aware data whitening technique not only ensures a direct mapping between singular values and compression loss, but is also able to achieve the same theoretical optimal compression loss as DRONE~\citep{DBLP:conf/nips/ChenYDH21}.
%
%
However, during SVD compression, DRONE stores all input activations in memory, which poses a challenge for LLM compression due to high memory consumption.
%
In contrast, \sysname incrementally updates its $XX^T$ matrix by adding the $xx^T$ of each new input $x$,  which is considerably smaller than the full input activation.
In~\cref{appendix:comparison_with_drone}, we provide our proof on \sysname achieving the same theoretical optimal compression loss as DRONE and discuss the advantages of \sysname over DRONE in memory saving, compression speed, and numerical stability in details.

\subsection{Parameter Update with Sequential Low-rank Approximation}
\label{subsec:lora_finetunig}
\textbf{Motivation:} 
Although the proposed truncation-aware data whitening technique helps preserve the accuracy during compression, as the compression ratio increases, the accuracy of the compressed LLM may degrade given that more larger singular values will be truncated by SVD compression.
Therefore, it is necessary to update the remaining parameters in the compressed LLM.

\textbf{Key Design:} 
%
\sysname proposes a variant of LoRA fine-tuning to update the remaining weight parameters of the compressed LLM for accuracy recovery. 
Specifically, given the two low-ranking matrices $W'_u, W'_v$ generated by truncation-aware data whitening,
instead of directly applying LoRA fine-tuning to the compressed weight matrix $W' = W'_u\times W'_v$ as standard LoRA does, we propose to apply LoRA on top of $W'_u$ and $W'_v$ separately to preserve their low-rank structures as follows:
\begin{align}
    W'_u \leftarrow W'_u + B_uA_u, W'_v\leftarrow W'_v + B_vA_v
\end{align}
where $A_u$, $B_u$, $A_v$, and $B_v$ are matrices used for LoRA fine-tuning. 

Simultaneously fine-tuning $W'_u$ and $W'_v$ will not guarantee a decrease in fine-tuning loss. This is because the derivatives of $W'_u$ and $W'_v$ are interdependent during the fine-tuning process, where optimization of one matrix may interfere with the optimization of the other. 
%
Therefore, as shown in Figure~\ref{fig:framework}, we propose a sequential low-rank approximation strategy to fine-tune $W'_u$ and $W'_v$ in a sequential manner. 
Specifically, we first freeze matrix $W'_v$ and fine-tune $W'_u$ with LoRA. We then perform the second-round LoRA fine-tuning on matrix $W'_v$ while freezing the updated weight matrix $W'_u$. Finally, we add $B_u\times A_u$ and $B_v\times A_v$ matrices to $W'_u$ and $W'_v$ to obtain the final compressed weight matrices.

\section{Experiments and Analysis}
\label{sec:experiments}
\vspace{-1mm}

\textbf{Baselines.}
We compare \sysname against three groups of methods: (1) Vanilla SVD and state-of-the-art SVD-based LLM compression methods: FWSVD~\citep{DBLP:conf/iclr/HsuHCLSJ22}, ASVD~\citep{DBLP:journals/corr/abs-2312-05821} (Section~\ref{subsec:overall_performance}) and FLAP (\cref{appendix:flap_comparison}).
(2) Other types of LLM compression methods. These include three state-of-the-art pruning-based LLM compression methods: LLM-Pruner~\citep{DBLP:conf/nips/MaFW23}, SliceGPT~\citep{DBLP:conf/iclr/AshkboosCNHH24}, and BlockPruner~\citep{DBLP:journals/corr/abs-2406-10594}, and three state-of-the-art quantization-based LLM compression methods: PB-LLM~\citep{DBLP:conf/iclr/YuanSD24}, BiLLM~\citep{DBLP:conf/icml/HuangLQLZ0M024}, and OneBit~\citep{DBLP:conf/nips/Xu0YWZLLC24} (Section~\ref{subsection:Comparison_with_other_compression_methods}). 
(3) Smaller LLM StableLM-3B~\citep{StableLM-3B-4E1T} pre-trained from scratch (\cref{appendix:pretrained_comparison}).

\textbf{Models and Datasets.} 
To demonstrate the generability of \sysname, we evaluate the performance of \sysname on seven models from three different LLM families at three different scales (LLaMA-7B, 13B, 30B, LLaMA2-7B~\citep{DBLP:journals/corr/abs-2307-09288}, OPT-6.7B~\citep{DBLP:journals/corr/abs-2205-01068}, Vicuna-7B~\citep{vicuna2023} and Mistral-7B~\citep{DBLP:journals/corr/abs-2310-06825}) and $10$ datasets including two language modeling datasets (WikiText-2~\citep{DBLP:conf/iclr/MerityX0S17}, and C4~\citep{DBLP:journals/jmlr/RaffelSRLNMZLL20}), six classification datasets (OpenbookQA~\citep{DBLP:conf/emnlp/MihaylovCKS18}, WinoGrande~\citep{DBLP:conf/aaai/SakaguchiBBC20}, HellaSwag~\citep{DBLP:conf/acl/ZellersHBFC19}, Arc\_e~\citep{DBLP:journals/corr/abs-1803-05457}, PIQA~\citep{DBLP:conf/aaai/BiskZLGC20}, MathQA~\citep{DBLP:conf/naacl/AminiGLKCH19}),  and two generation datasets (TruthfulQA~\citep{DBLP:conf/acl/LinHE22} and GSM8K~\citep{DBLP:journals/corr/abs-2110-14168}) with the LM-Evaluation-Harness framework~\citep{eval-harness}.

\textbf{Implementation Details.}
To ensure a fair comparison, we followed ASVD~\citep{DBLP:journals/corr/abs-2312-05821} to randomly select 256 samples from WikiText-2 as the calibration data. 
We followed the same configuration used in LLM-Pruner~\citep{DBLP:conf/nips/MaFW23} to use Alpaca~\citep{alpaca} dataset with 50K samples for parameter update in \sysname.
The inference efficiency experiment is conducted on both NVIDIA A100 GPU and AMD EPYC 7643 CPU while the other experiments are conducted on NVIDIA A100 GPUs.

\vspace{-2mm}
\subsection{Comparison with State-of-the-Art SVD-based LLM compression Methods}
\label{subsec:overall_performance}
\vspace{-2mm}
First, we compare the performance of \sysname with state-of-the-art SVD-based LLM compression methods from four aspects: (1) performance under different compression ratios, (2) performance on different LLMs, (3) performance on LLMs with larger scales, and (4) compression speed (\cref{appendix:compression_speed}). 
Given that all the SVD-based baselines do not incorporate LoRa fine-tuning, to ensure a fair comparison, we also compare to 
\sysname with truncation-aware data whitening only (denoted as \sysname \texttt{(W)}).

\accuracyTable
\differntllmaccuracyTable
\textbf{Performance under Different Compression Ratios.}
We first evaluate the performance of LLaMA-7B compressed by \sysname, \sysname \texttt{(W)}, and the SVD-based baselines under compression ratios ranging from 20\% to 80\% on all $10$ datasets. 
The results are summarized in\cref{tab:dataset_acc}. Both \sysname and \sysname \texttt{(W)} consistently outperform vanilla SVD, FWSVD and ASVD across all the compression ratios. In particular, when the compression ratio is 40\% and above, \sysname reduces the perplexity by more than 99\% on two language modeling datasets and achieves over 400\% higher average accuracy on six classification datasets. More importantly, the results on two generation datasets (TruthfulQA, GSM8K) of all three baselines when compression ratios are 60\% and above are zero, meaning that the compressed LLMs totally lose their generation ability. In contrast, \sysname still outputs good generation even under 80\% compression ratio. 
Example contents generated by the compressed LLMs are included in~\cref{sec:appendix_example}.
%


\begin{wrapfigure}{r}{0.48\textwidth} 
  \centering
  \vspace{-5mm}
  \largellmaccuracyTable
\end{wrapfigure}

\textbf{Performance on Different LLMs.} To examine the generability of \sysname and \sysname \texttt{(W)} across different LLMs, we compare the performance between \sysname and the baselines on four different models from three different LLM families -- OPT-6.7B (OPT family), LLaMA 2-7B (LLaMA family), Mistral-7B (Mistral family), and Vicuna-7B (LLaMA family)  -- under 20\% compression ratio on WikiText-2 and six classification datasets. 
As shown in~\cref{tab:different_llm_acc}, both \sysname and \sysname \texttt{(W)} consistently outperform baselines on all four LLMs, and exhibits more stable performance across different LLMs, especially compared to vanilla SVD and FWSVD.

\textbf{Performance on LLMs with Larger Scales.} To examine the generability of \sysname and \sysname \texttt{(W)} on LLMs with larger scales, we compare the performance between \sysname and the baselines on LLaMA-13B and LLaMA-30B under 20\% compression ratio. As shown in~\cref{tab:large_llm_acc}, both \sysname and \sysname \texttt{(W)} consistently outperform vanilla SVD, FWSVD, and ASVD on both model sizes.

\vspace{-1mm}
\subsection{Inference Efficiency of \sysname }
\label{subsec:inference_speedup}
\vspace{-1mm}

\textbf{Theoretical Analysis of Inference Efficiency.} Assume \sysname compresses the weight matrix {\small$W \in \mathbb{R}^{d \times n}$} into two low-ranking matrices {\small$W_u' \in \mathbb{R}^{d \times r}, W_v' \in \mathbb{R}^{r \times n}$}. The compression ratio is then calculated as {\small$R_w = 1-\frac{(d+n) r}{d n}$}.

\underline{(1) Compute Complexity Analysis:} Given input {\small$X \in \mathbb{R}^{n \times d}$}, instead of recalculating the full weight matrix {\small$W^{\prime}=W_u' \times W_v'$} and then computing the output {\small$W^{\prime} \times X$}, \sysname calculates the intermediate state {\small$M=W_v' \times X$} and then computes the output {\small$Y=W_u' \times M$}. 
In this way, the computation complexity is reduced from {\small$O\left(d^2 n\right)$} to {\small$O\left(d^2 r+r n d\right)$}. 
Taking compression ratio {\small$R_w=50\%$} as an example, since {\small$R_w=1 - \frac{(d+n) r}{d n}$}, we have {\small$r=\frac{d n}{2(d+n)}$}. Then the computation complexity is
{\small$O\left(d^2 r+r n d\right)=O(r d(d+n))=O\left(\frac{d^2 n}{2}\right)=\frac{1}{2} O\left(d^2 n\right)$}, which reduces {\small$50\%$}. In general, given any compression ratio $R_w$, the computation complexity is reduced to {\small$1 - R_w$} times of the original.

\begin{figure}[htbp]
 \vspace{-0mm}
     \begin{minipage}{0.98\textwidth}
     \centering
    \includegraphics[width=1\linewidth]{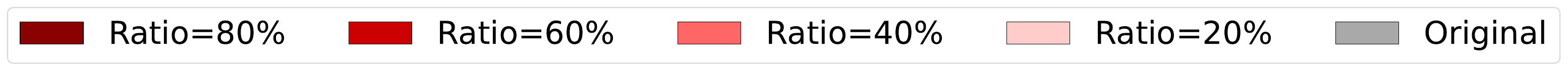}
     \end{minipage}
     \\
     \begin{minipage}{1\textwidth}
        \subfigure[Varying Batch Size on GPU]{
            \centering
            \includegraphics[width=0.48\textwidth]{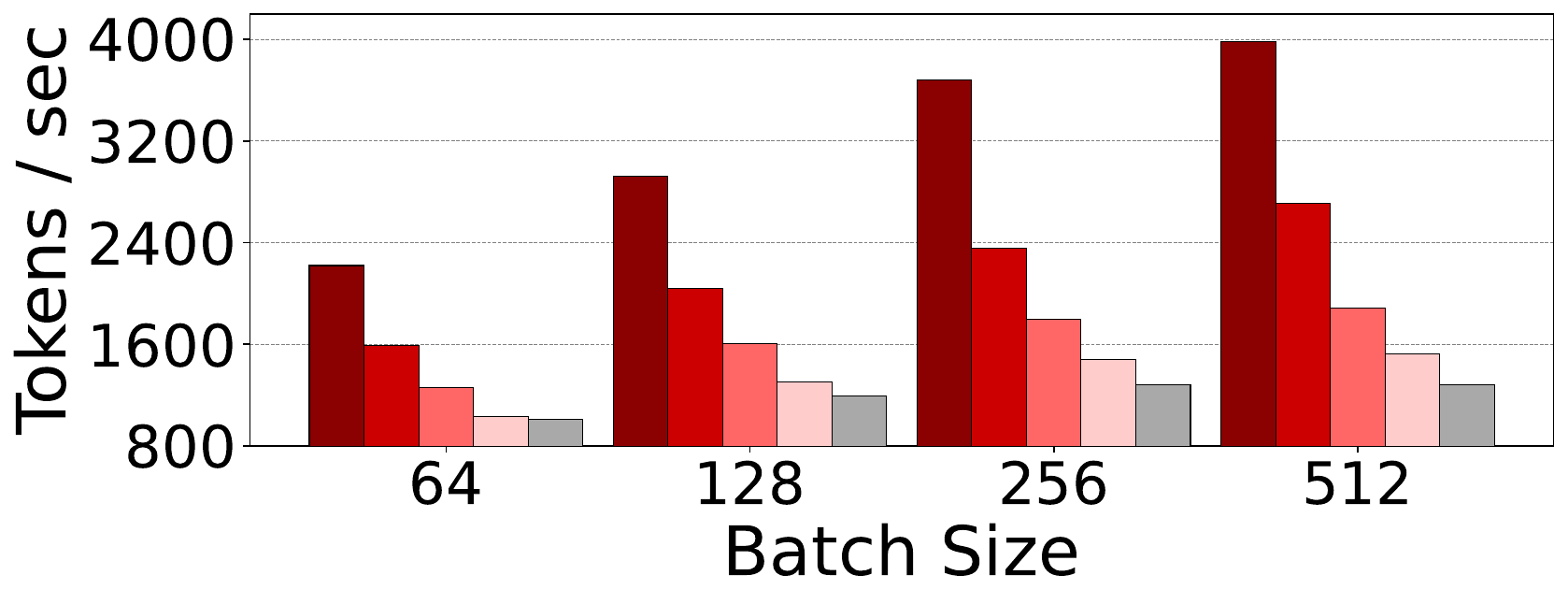}
            \label{fig:bs_gpu}
            }
        \subfigure[Varying Sequence Length on GPU]{
            \centering
            \includegraphics[width=0.48\textwidth]{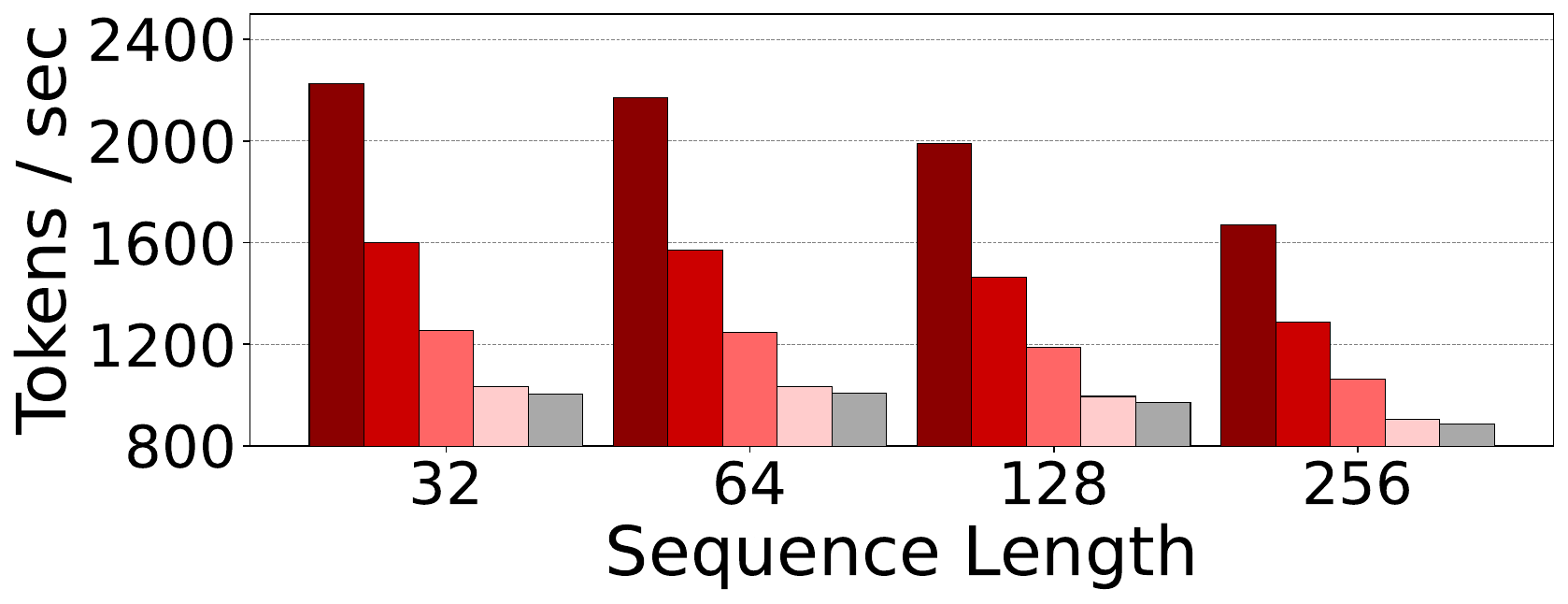}
            \label{fig:sl_gpu}
            }
        \\
        \subfigure[Varying Batch Size on CPU]{
            \centering
            \includegraphics[width=0.48\textwidth]{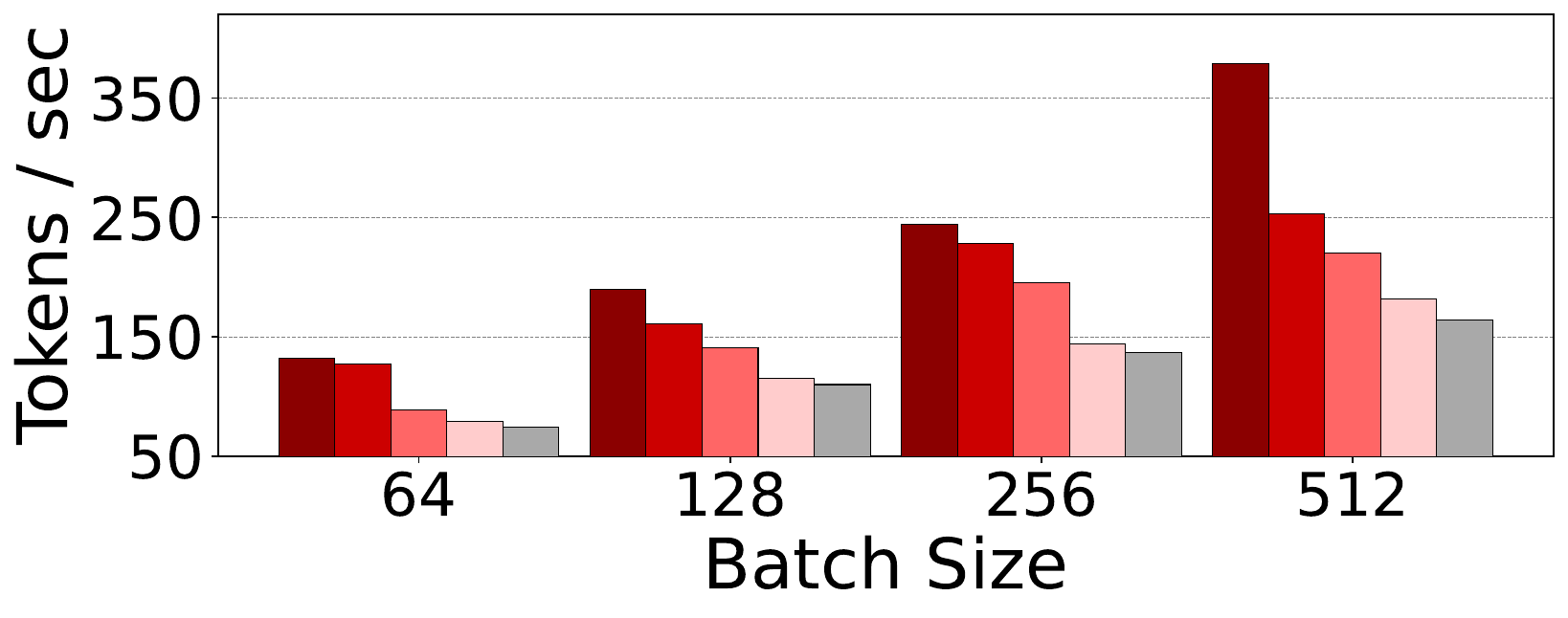}
            \label{fig:bs_cpu}
            }
        \subfigure[Varying Sequence Length on CPU]{
            \centering
            \includegraphics[width=0.48\textwidth]{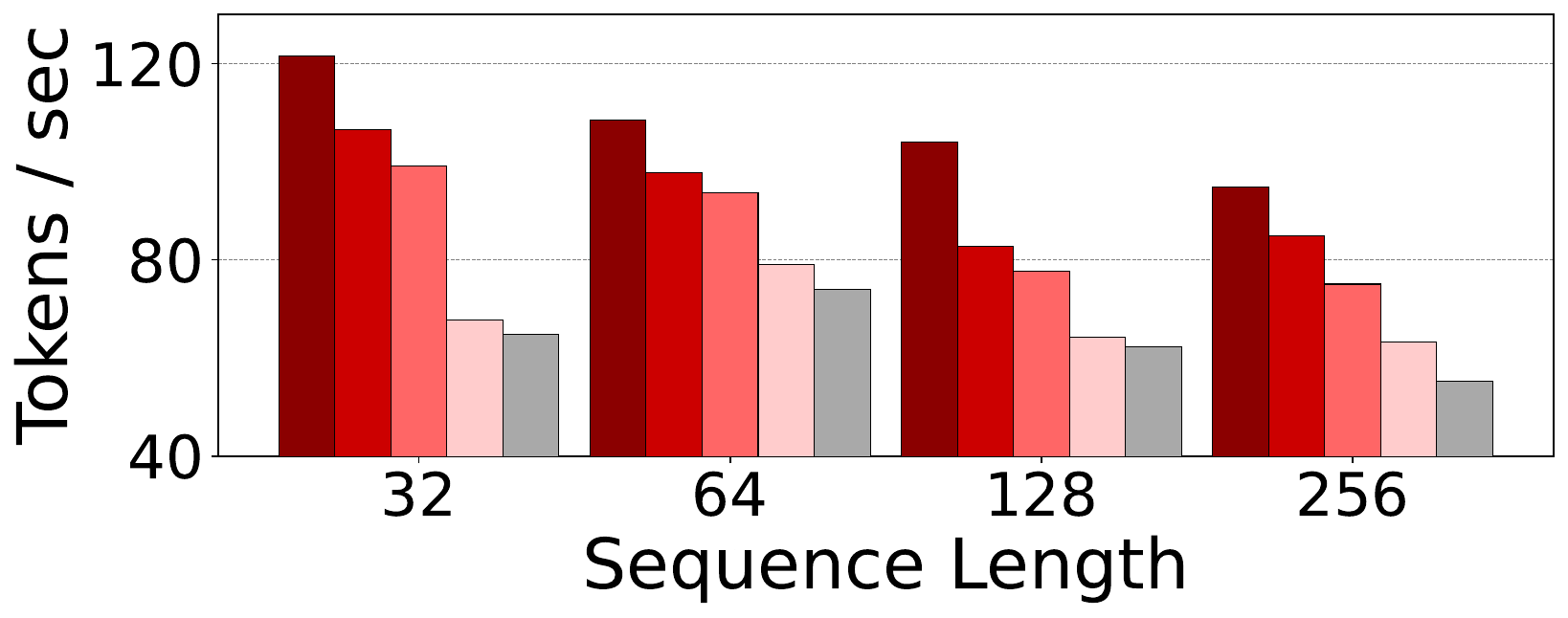}
            \label{fig:sl_cpu}
            }
        \DeclareGraphicsExtensions.
        \vspace{-2mm}
        \caption{Throughput (tokens/sec) of LLaMA-7B and its compressed version by \sysname under different compression ratios on a single A100 GPU under different batch sizes (a) and different sequence lengths (b) and on a single AMD EPYC 7643 CPU under different batch sizes (c) and different sequence lengths (d). For (a) and (c), sequence length is 32. For (b) and (d), batch size is 64.}
        \vspace{-2mm}
        \label{fig:gpu_throughput}
    \end{minipage}
\end{figure}

\underline{(2) Inference Memory Analysis:} 
Since \sysname does not recalculate the full weight $W^{\prime}=W_u' \times W_v'$, the weight memory is reduced to $1 - R_w$ times of the original during inference.
As another advantage, \sysname is able to reduce the runtime KV cache memory as well~\citep{DBLP:journals/corr/abs-2406-13035, wan2025meda,DBLP:journals/corr/abs-2409-09808}. Specifically, instead of keeping {\small$W_u' \times W_v' \times X$} in the KV cache, \sysname provides the option to store the intermediate result {\small$M=W_v' \times X$} in the KV cache and recomputes the original key and value states with the decomposed weight matrix {\small$W_u'$} if required. As such, the runtime KV cache is reduced to {\small$\frac{r}{d}= (1 - R_w)\times \frac{d}{n+d}$} times of the original. Moreover, since {\small$W_u'$} is already stored as the weight matrix in the decomposed LLM, the original intermediate state matrix can still be recovered by {\small$W_u'\times M$} without accuracy drop. Therefore, \sysname provides a unified solution that enables \textit{simultaneous} model compression and KV cache compression.

\textbf{Inference Speedup on Hardware.} 
To quantify inference speedup achieved by \sysname on real hardware, we measure the numbers of tokens that LLaMA-7B and its compressed version by \sysname generate per second (i.e., throughput) under different batch sizes and sequence lengths on a single NVIDIA A100 GPU and a single AMD EPYC 7643 CPU, respectively. 
The results are shown in~\cref{fig:gpu_throughput}. We have three observations. 
(1) Under a specific batch size or sequence length, the speedup achieved by \sysname related to the original model increases as the compression ratio increases.
%
(2) Under a specific compression ratio, the speedup  achieved by \sysname related to the original model becomes more significant as the batch size increases or as the sequence length decreases.
(3) The above two observations are valid for both GPU and CPU.

\begin{figure}[htbp]
\vspace{-5mm}
    \centering
    \begin{minipage}{0.5\textwidth}
    \centering
    \vspace{4mm}
        \centering
        \includegraphics[width=1\textwidth]{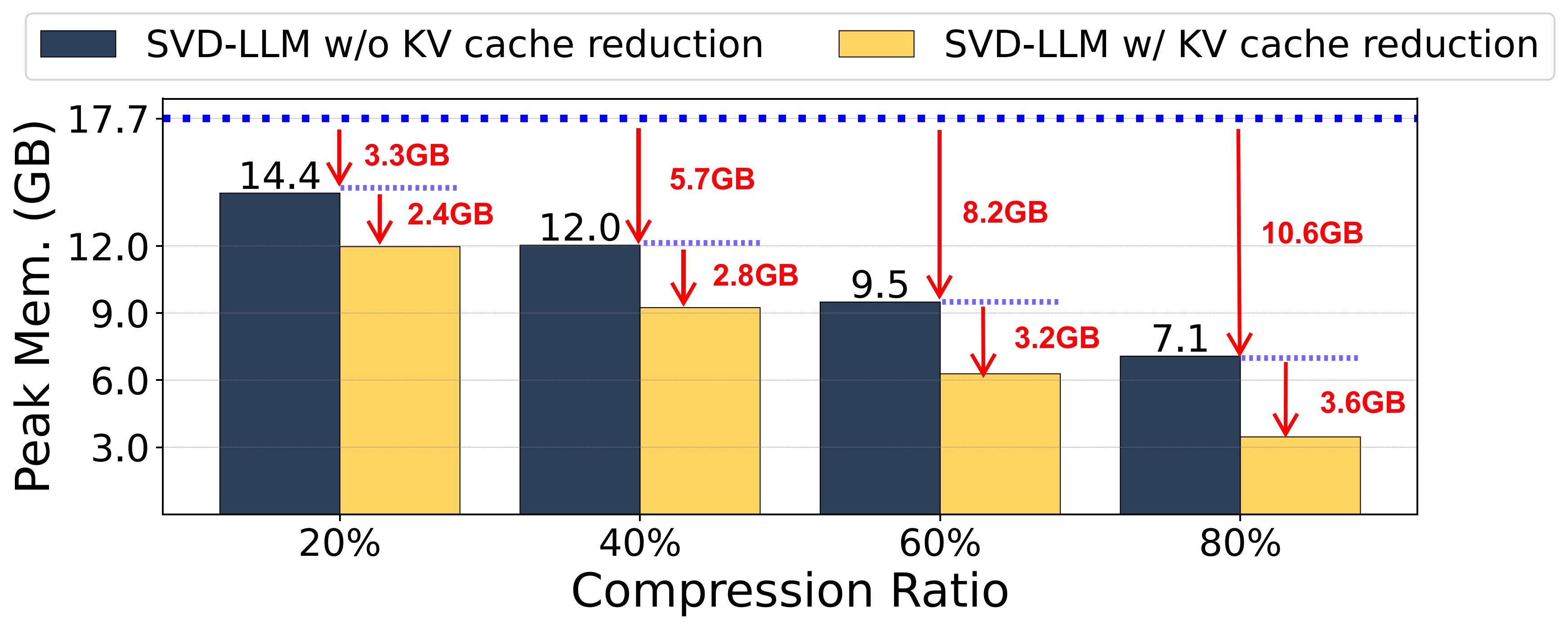}
        \vspace{-5mm}
        \caption{Peak memory to generate 128 tokens with batch size of 32 using LLaMA-7B compressed by \sysname w/ and w/o KV-cache reduction. The dotted line indicates the peak memory of the original LLaMA-7B. The memory reduction from the dotted line to the blue bars mainly comes from the model compression. The memory reduction from the blue to the yellow bars mainly comes from the reduced footprint of the KV cache.}
        \label{fig:kv}
    \end{minipage}%
    \hspace{0.1cm}
    \begin{minipage}{0.485\textwidth}
    \centering
    \sensitivityTable
\end{minipage}
\vspace{-3mm}
\end{figure}

\textbf{Inference Memory Reduction on Hardware.} 
Lastly, we evaluate the inference memory saving, including both weight memory and runtime KV cache memory saving on a single A100 GPU. 
Specifically, we measure the peak memory footprint during inference when generating 128 tokens with batch size of 32 using LLaMA-7B compressed by \sysname under different compression ratios w/ and w/o considering KV cache reduction. 
The results are illustrated in~\cref{fig:kv} where the memory reduction from the dotted line to the blue bars comes mainly from model weight compression and the memory reduction from the blue bars to the yellow bars comes mainly from KV cache compression. 
As shown, both weight memory saving and runtime KV cache memory saving brought by \sysname are near linear to the compression ratio.

\vspace{-1mm}
\subsection{Ablation Study}
\label{subsec:ablation_study}
\vspace{-1mm}

%

\textbf{Modular Sensitivity Study.}
We conduct ablation studies to evaluate the separate contributions of the two key components (i.e., truncation-aware data whitening and parameter update with sequential low-rank approximation) of \sysname. 
Let \sysname \texttt{(W)} denote the version of \sysname with truncation-aware data whitening only; \sysname \texttt{(U)} denote the version of \sysname with \xin{normal SVD truncation} and parameter update with sequential low-rank approximation.
As shown in~\cref{tab:sensitivity}, we have three observations. 
(1) \sysname\texttt{(W)}, \sysname\texttt{(U)} and \sysname  consistently outperform ASVD across all the compression ratios. Notably, when the compression ratio is at and above 40\%, all of them reduce the perplexity by more than 99\% compared to ASVD. 
(2) \sysname consistently outperforms both \sysname\texttt{(U)} and \sysname\texttt{(W)} across all compression ratios.
This result demonstrates the unique contribution from each of the two key components and the importance of combining both components to achieve the best performance.
(3) Comparing between \sysname\texttt{(W)} and \sysname\texttt{(U)}, \sysname\texttt{(W)} achieves a lower perplexity compared to \sysname\texttt{(U)} across all compression ratios. This result indicates that truncation-aware data whitening plays a more significant role than parameter update with sequential
low-rank approximation.

\begin{figure}[htbp]
\vspace{-4mm}
    \centering
    \begin{minipage}{0.5\textwidth}
    \vspace{2mm}
    \centering
        \subfigure[Change of Number]{
        \centering
        \includegraphics[width=0.455\textwidth]{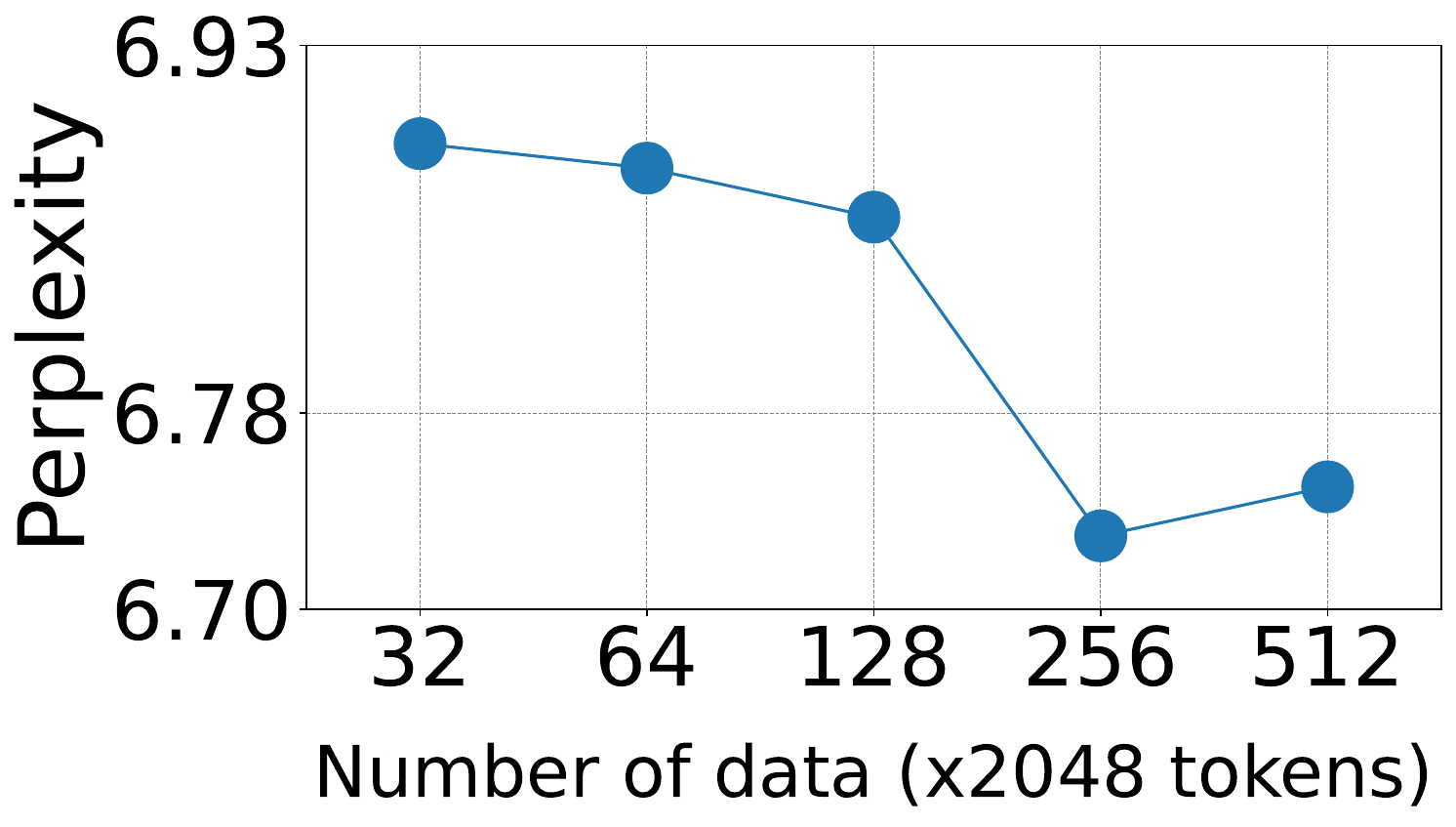}
        \label{fig:cali_size}
        }
        \subfigure[Change of Seed]{
        \centering
        \includegraphics[width=0.45\textwidth]{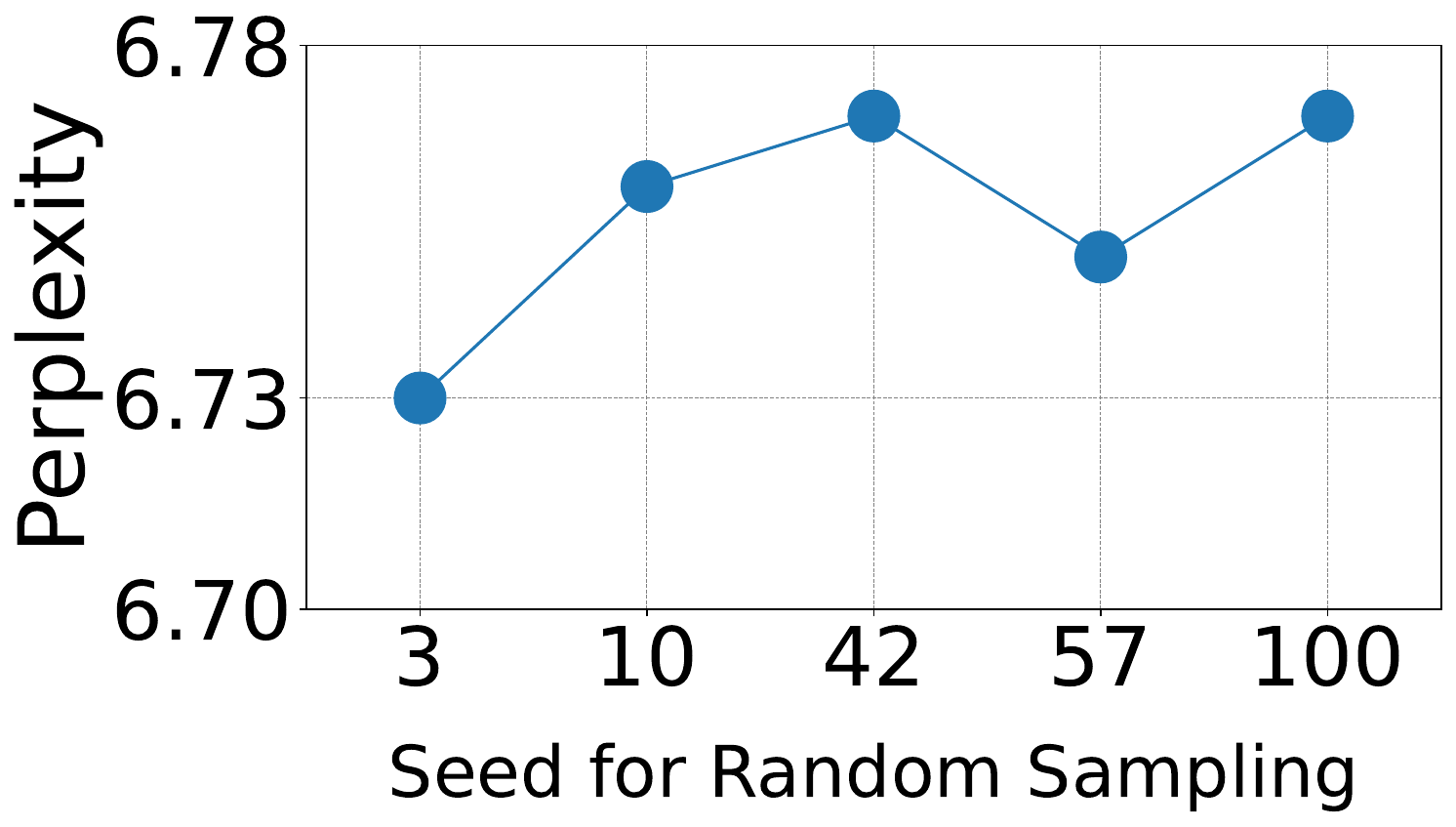}
        \label{fig:seed}
        }
        \vspace{-3mm}
        \DeclareGraphicsExtensions.
        \caption{Perplexity of LLaMA-7B under 20\% compression ratio using calibration data sampled with different number or  seeds from WikiText-2.}
        \label{fig:cali}
        \vspace{6mm}
        \loraTable
    \end{minipage}%
    \hspace{0.1cm}
    \begin{minipage}{0.485\textwidth}
    \vspace{3.5mm}
    \centering
    \datasetTable
\end{minipage}
\vspace{-3mm}
\end{figure}

\textbf{Impact of Calibration Data.}
Next, we examine the impact of calibration data on \sysname. 
\cref{fig:cali} and~\cref{tab:different_ds_acc} summarize the performance of compressed LLaMA-7B when changing three key characteristics of the calibration data: (1) number of calibration data, (2) the seed used to randomly sample the calibration data, and (3)  dataset from which calibration data is sampled.
As shown, the changes of the three key characteristics on calibration data incur no more than 3\% to the final performance, indicating that the sensitivity of \sysname on calibration data is limited.

\textbf{Impact of Updating Order.}
Lastly, we examine the impact of updating order in the parameter update with sequential low-rank approximation component to the final performance of the compressed LLM. 
Table~\ref{tab:lora_acc} shows the performance of compressed LLaMA-7B under 20\% to 80\% compression ratios on WikiText-2 with different updating orders. As shown, there is only a small difference of the final performance between updating matrix $W'_u$ first and updating matrix $W'_v$ first. This result indicates \sysname is not sensitive to the updating order.

More ablation studies are included in~\cref{appendix:more_ablation}.

\begin{figure}[t]
    \centering
    \begin{minipage}{.52\textwidth}
        \svdpruneTable
    \end{minipage}
    \hspace{0.1cm}
    \begin{minipage}{.45\textwidth}
      \svdquantTable
    \end{minipage}%
\end{figure}

\subsection{Comparison with Other Types of LLM Compression Methods}
\label{subsection:Comparison_with_other_compression_methods}
\vspace{-1mm}
\sysname is orthogonal to other post-training LLM compression methods including pruning and quantization~\citep{DBLP:journals/tmlr/Wan0LA0LQYZZC024, shen2025efficient}. Lastly, we compare the performance of \sysname with state-of-the-art structured pruning-based and quantization-based LLM compression methods. As discussed in Section~\ref{sec:related_works}, since unstructured pruning methods are difficult to achieve its efficiency on hardware, we do not make a comparison with them in this experiment.

\vspace{-0mm}
\textbf{Comparison with Structured Pruning.}
First, we compare \sysname with three state-of-the-art structured pruning-based LLM compression methods: LLM-Pruner~\citep{DBLP:conf/nips/MaFW23}, SliceGPT~\citep{DBLP:conf/iclr/AshkboosCNHH24} and BlockPruner~\citep{DBLP:journals/corr/abs-2406-10594} under the same memory budget, ranging from 10 GB to 7 GB on LLaMA-7B using WikiText-2 dataset. 
As shown in Table~\ref{tab:svd_prune}, \sysname outperforms all three structured pruning-based LLM compression methods. In particular, \sysname achieves up to 56\% reduction in perplexity under 7G memory budget.

\vspace{-0mm}
\textbf{Comparison with Quantization.} Finally, we compare \sysname with three state-of-the-art quantization-based LLM compression methods: BiLLM~\citep{DBLP:conf/icml/HuangLQLZ0M024}, PB-LLM~\citep{DBLP:conf/iclr/YuanSD24}, and OneBit~\citep{DBLP:conf/nips/Xu0YWZLLC24}, which push the frontier to 1-bit quantization. 
Among them, both BiLLM and PB-LLM are post-training methods, and OneBit is training-required. 
The results on LLaMA-7B using WikiText-2 dataset are shown in Table~\ref{tab:svd_quant}. 
We have three observations. 
(1) Compared to post-training methods PB-LLM and BiLLM, \sysname achieves the best performance. 
(2) Training-required method OneBit outperforms \sysname. This result is expected because OneBit involves resource-intensive retraining using large-scale datasets to boost the accuracy after compression. 
(3) Lastly, we combine \sysname with a 2-bit quantization-based post-training LLM compression method QuIP\#~\citep{DBLP:conf/icml/TsengCSKS24}. This is achieved by first applying \sysname to the LLM under 40\% compression ratio, and then applying QuIP\# for 2-bit quantization on the compressed model ($W'_u$ and $W'_v$) generated from \sysname. As shown in Table~\ref{tab:svd_quant}, \sysname outperforms state-of-the-art 1-bit training-required method OneBit \textit{without} involving resource-intensive retraining. This result demonstrates the potential of a hybrid approach -- integrating SVD-based and quantization-based compression techniques -- to push the boundaries of post-training LLM compression.

%


\vspace{-2mm}
\section{Conclusion}
\vspace{-2mm}
\label{sec:conclusion}
In this work, we present \sysname, a SVD-based
post-training LLM compression method. 
\sysname proposes a truncation-aware data whitening technique to guide which singular values to be truncated with minimal compression loss. It also introduces a sequential low-rank approximation strategy to compensate for accuracy degradation caused by singular value truncation.
We evaluate \sysname on $10$ datasets and seven models from three LLM families at three scales. Our results demonstrate
 the superiority of SVD-LLM over state-of-the-arts, especially at high model compression ratios.

\vspace{-2mm}
\section{Acknowledgement}
\vspace{-2mm}

This work is supported in part by NSF Award NeTS-2312675.

\bibliography{Reference}

\begin{thebibliography}{43}
\providecommand{\natexlab}[1]{#1}
\providecommand{\url}[1]{\texttt{#1}}
\expandafter\ifx\csname urlstyle\endcsname\relax
  \providecommand{\doi}[1]{doi: #1}\else
  \providecommand{\doi}{doi: \begingroup \urlstyle{rm}\Url}\fi

\bibitem[Amini et~al.(2019)Amini, Gabriel, Lin, Koncel{-}Kedziorski, Choi, and Hajishirzi]{DBLP:conf/naacl/AminiGLKCH19}
Aida Amini, Saadia Gabriel, Shanchuan Lin, Rik Koncel{-}Kedziorski, Yejin Choi, and Hannaneh Hajishirzi.
\newblock Mathqa: Towards interpretable math word problem solving with operation-based formalisms.
\newblock In \emph{{NAACL-HLT} {(1)}}, pages 2357--2367. Association for Computational Linguistics, 2019.

\bibitem[An et~al.(2024)An, Zhao, Yu, Tang, and Wang]{DBLP:conf/aaai/AnZYTW24}
Yongqi An, Xu~Zhao, Tao Yu, Ming Tang, and Jinqiao Wang.
\newblock Fluctuation-based adaptive structured pruning for large language models.
\newblock In \emph{{AAAI}}, pages 10865--10873. {AAAI} Press, 2024.

\bibitem[Ashkboos et~al.(2024)Ashkboos, Croci, Nascimento, Hoefler, and Hensman]{DBLP:conf/iclr/AshkboosCNHH24}
Saleh Ashkboos, Maximilian~L. Croci, Marcelo Gennari~Do Nascimento, Torsten Hoefler, and James Hensman.
\newblock Slicegpt: Compress large language models by deleting rows and columns.
\newblock In \emph{{ICLR}}. OpenReview.net, 2024.

\bibitem[Bisk et~al.(2020)Bisk, Zellers, Bras, Gao, and Choi]{DBLP:conf/aaai/BiskZLGC20}
Yonatan Bisk, Rowan Zellers, Ronan~Le Bras, Jianfeng Gao, and Yejin Choi.
\newblock {PIQA:} reasoning about physical commonsense in natural language.
\newblock In \emph{{AAAI}}, pages 7432--7439. {AAAI} Press, 2020.

\bibitem[Chen et~al.(2021)Chen, Yu, Dhillon, and Hsieh]{DBLP:conf/nips/ChenYDH21}
Patrick~H. Chen, Hsiang{-}Fu Yu, Inderjit~S. Dhillon, and Cho{-}Jui Hsieh.
\newblock {DRONE:} data-aware low-rank compression for large {NLP} models.
\newblock In \emph{NeurIPS}, pages 29321--29334, 2021.

\bibitem[Chiang et~al.(2023)Chiang, Li, Lin, Sheng, Wu, Zhang, Zheng, Zhuang, Zhuang, Gonzalez, Stoica, and Xing]{vicuna2023}
Wei-Lin Chiang, Zhuohan Li, Zi~Lin, Ying Sheng, Zhanghao Wu, Hao Zhang, Lianmin Zheng, Siyuan Zhuang, Yonghao Zhuang, Joseph~E. Gonzalez, Ion Stoica, and Eric~P. Xing.
\newblock Vicuna: An open-source chatbot impressing gpt-4 with 90\%* chatgpt quality, March 2023.
\newblock URL \url{https://lmsys.org/blog/2023-03-30-vicuna/}.

\bibitem[Clark et~al.(2018)Clark, Cowhey, Etzioni, Khot, Sabharwal, Schoenick, and Tafjord]{DBLP:journals/corr/abs-1803-05457}
Peter Clark, Isaac Cowhey, Oren Etzioni, Tushar Khot, Ashish Sabharwal, Carissa Schoenick, and Oyvind Tafjord.
\newblock Think you have solved question answering? try arc, the {AI2} reasoning challenge.
\newblock \emph{CoRR}, abs/1803.05457, 2018.

\bibitem[Cobbe et~al.(2021)Cobbe, Kosaraju, Bavarian, Chen, Jun, Kaiser, Plappert, Tworek, Hilton, Nakano, Hesse, and Schulman]{DBLP:journals/corr/abs-2110-14168}
Karl Cobbe, Vineet Kosaraju, Mohammad Bavarian, Mark Chen, Heewoo Jun, Lukasz Kaiser, Matthias Plappert, Jerry Tworek, Jacob Hilton, Reiichiro Nakano, Christopher Hesse, and John Schulman.
\newblock Training verifiers to solve math word problems.
\newblock \emph{CoRR}, abs/2110.14168, 2021.

\bibitem[Frantar and Alistarh(2023)]{DBLP:conf/icml/FrantarA23}
Elias Frantar and Dan Alistarh.
\newblock Sparsegpt: Massive language models can be accurately pruned in one-shot.
\newblock In \emph{{ICML}}, volume 202 of \emph{Proceedings of Machine Learning Research}, pages 10323--10337. {PMLR}, 2023.

\bibitem[Gao et~al.(2023)Gao, Tow, Abbasi, Biderman, Black, DiPofi, Foster, Golding, Hsu, Le~Noac'h, Li, McDonell, Muennighoff, Ociepa, Phang, Reynolds, Schoelkopf, Skowron, Sutawika, Tang, Thite, Wang, Wang, and Zou]{eval-harness}
Leo Gao, Jonathan Tow, Baber Abbasi, Stella Biderman, Sid Black, Anthony DiPofi, Charles Foster, Laurence Golding, Jeffrey Hsu, Alain Le~Noac'h, Haonan Li, Kyle McDonell, Niklas Muennighoff, Chris Ociepa, Jason Phang, Laria Reynolds, Hailey Schoelkopf, Aviya Skowron, Lintang Sutawika, Eric Tang, Anish Thite, Ben Wang, Kevin Wang, and Andy Zou.
\newblock A framework for few-shot language model evaluation, 12 2023.
\newblock URL \url{https://zenodo.org/records/10256836}.

\bibitem[Golub et~al.(1987)Golub, Hoffman, and Stewart]{GOLUB1987317}
G.H. Golub, Alan Hoffman, and G.W. Stewart.
\newblock A generalization of the eckart-young-mirsky matrix approximation theorem.
\newblock \emph{Linear Algebra and its Applications}, 88-89:\penalty0 317--327, 1987.
\newblock ISSN 0024-3795.
\newblock \doi{https://doi.org/10.1016/0024-3795(87)90114-5}.
\newblock URL \url{https://www.sciencedirect.com/science/article/pii/0024379587901145}.

\bibitem[Gozalo{-}Brizuela and Garrido{-}Merch{\'{a}}n(2023)]{DBLP:journals/corr/abs-2306-02781}
Roberto Gozalo{-}Brizuela and Eduardo~C. Garrido{-}Merch{\'{a}}n.
\newblock A survey of generative {AI} applications.
\newblock \emph{CoRR}, abs/2306.02781, 2023.

\bibitem[Hsu et~al.(2022)Hsu, Hua, Chang, Lou, Shen, and Jin]{DBLP:conf/iclr/HsuHCLSJ22}
Yen{-}Chang Hsu, Ting Hua, Sungen Chang, Qian Lou, Yilin Shen, and Hongxia Jin.
\newblock Language model compression with weighted low-rank factorization.
\newblock In \emph{{ICLR}}. OpenReview.net, 2022.

\bibitem[Huang et~al.(2024)Huang, Liu, Qin, Li, Zhang, Liu, Magno, and Qi]{DBLP:conf/icml/HuangLQLZ0M024}
Wei Huang, Yangdong Liu, Haotong Qin, Ying Li, Shiming Zhang, Xianglong Liu, Michele Magno, and Xiaojuan Qi.
\newblock Billm: Pushing the limit of post-training quantization for llms.
\newblock In \emph{{ICML}}. OpenReview.net, 2024.

\bibitem[Jiang et~al.(2023)Jiang, Sablayrolles, Mensch, Bamford, Chaplot, de~Las~Casas, Bressand, Lengyel, Lample, Saulnier, Lavaud, Lachaux, Stock, Scao, Lavril, Wang, Lacroix, and Sayed]{DBLP:journals/corr/abs-2310-06825}
Albert~Q. Jiang, Alexandre Sablayrolles, Arthur Mensch, Chris Bamford, Devendra~Singh Chaplot, Diego de~Las~Casas, Florian Bressand, Gianna Lengyel, Guillaume Lample, Lucile Saulnier, L{\'{e}}lio~Renard Lavaud, Marie{-}Anne Lachaux, Pierre Stock, Teven~Le Scao, Thibaut Lavril, Thomas Wang, Timoth{\'{e}}e Lacroix, and William~El Sayed.
\newblock Mistral 7b.
\newblock \emph{CoRR}, abs/2310.06825, 2023.

\bibitem[Lin et~al.(2022)Lin, Hilton, and Evans]{DBLP:conf/acl/LinHE22}
Stephanie Lin, Jacob Hilton, and Owain Evans.
\newblock Truthfulqa: Measuring how models mimic human falsehoods.
\newblock In \emph{{ACL} {(1)}}, pages 3214--3252. Association for Computational Linguistics, 2022.

\bibitem[Lin et~al.(2024)Lin, Tang, Yang, Zhang, Xiao, Gan, and Han]{DBLP:journals/corr/abs-2405-04532}
Yujun Lin, Haotian Tang, Shang Yang, Zhekai Zhang, Guangxuan Xiao, Chuang Gan, and Song Han.
\newblock Qserve: {W4A8KV4} quantization and system co-design for efficient {LLM} serving.
\newblock \emph{CoRR}, abs/2405.04532, 2024.

\bibitem[Ma et~al.(2023)Ma, Fang, and Wang]{DBLP:conf/nips/MaFW23}
Xinyin Ma, Gongfan Fang, and Xinchao Wang.
\newblock Llm-pruner: On the structural pruning of large language models.
\newblock In \emph{NeurIPS}, 2023.

\bibitem[Merity et~al.(2017)Merity, Xiong, Bradbury, and Socher]{DBLP:conf/iclr/MerityX0S17}
Stephen Merity, Caiming Xiong, James Bradbury, and Richard Socher.
\newblock Pointer sentinel mixture models.
\newblock In \emph{{ICLR} (Poster)}. OpenReview.net, 2017.

\bibitem[Meyer(2000)]{DBLP:books/siam/Meyer00}
Carl~Dean Meyer.
\newblock \emph{Matrix Analysis and Applied Linear Algebra}.
\newblock {SIAM}, 2000.

\bibitem[Mihaylov et~al.(2018)Mihaylov, Clark, Khot, and Sabharwal]{DBLP:conf/emnlp/MihaylovCKS18}
Todor Mihaylov, Peter Clark, Tushar Khot, and Ashish Sabharwal.
\newblock Can a suit of armor conduct electricity? {A} new dataset for open book question answering.
\newblock In \emph{{EMNLP}}, pages 2381--2391. Association for Computational Linguistics, 2018.

\bibitem[Raffel et~al.(2020)Raffel, Shazeer, Roberts, Lee, Narang, Matena, Zhou, Li, and Liu]{DBLP:journals/jmlr/RaffelSRLNMZLL20}
Colin Raffel, Noam Shazeer, Adam Roberts, Katherine Lee, Sharan Narang, Michael Matena, Yanqi Zhou, Wei Li, and Peter~J. Liu.
\newblock Exploring the limits of transfer learning with a unified text-to-text transformer.
\newblock \emph{J. Mach. Learn. Res.}, 21:\penalty0 140:1--140:67, 2020.

\bibitem[Sakaguchi et~al.(2020)Sakaguchi, Bras, Bhagavatula, and Choi]{DBLP:conf/aaai/SakaguchiBBC20}
Keisuke Sakaguchi, Ronan~Le Bras, Chandra Bhagavatula, and Yejin Choi.
\newblock Winogrande: An adversarial winograd schema challenge at scale.
\newblock In \emph{{AAAI}}, pages 8732--8740. {AAAI} Press, 2020.

\bibitem[Shen et~al.(2024)Shen, Wan, Wang, and Zhang]{DBLP:journals/corr/abs-2409-09808}
Hui Shen, Zhongwei Wan, Xin Wang, and Mi~Zhang.
\newblock Famba-v: Fast vision mamba with cross-layer token fusion.
\newblock \emph{CoRR}, abs/2409.09808, 2024.

\bibitem[Shen et~al.(2025)Shen, Zhang, Xiong, Hu, Chen, Wan, Wang, Zhang, Gong, Bao, et~al.]{shen2025efficient}
Hui Shen, Jingxuan Zhang, Boning Xiong, Rui Hu, Shoufa Chen, Zhongwei Wan, Xin Wang, Yu~Zhang, Zixuan Gong, Guangyin Bao, et~al.
\newblock Efficient diffusion models: A survey.
\newblock \emph{arXiv preprint arXiv:2502.06805}, 2025.

\bibitem[Taori et~al.(2023)Taori, Gulrajani, Zhang, Dubois, Li, Guestrin, Liang, and Hashimoto]{alpaca}
Rohan Taori, Ishaan Gulrajani, Tianyi Zhang, Yann Dubois, Xuechen Li, Carlos Guestrin, Percy Liang, and Tatsunori~B. Hashimoto.
\newblock Stanford alpaca: An instruction-following llama model.
\newblock \url{https://github.com/tatsu-lab/stanford_alpaca}, 2023.

\bibitem[Touvron et~al.(2023)Touvron, Martin, Stone, Albert, Almahairi, Babaei, Bashlykov, Batra, Bhargava, Bhosale, Bikel, Blecher, Canton{-}Ferrer, Chen, Cucurull, Esiobu, Fernandes, Fu, Fu, Fuller, Gao, Goswami, Goyal, Hartshorn, Hosseini, Hou, Inan, Kardas, Kerkez, Khabsa, Kloumann, Korenev, Koura, Lachaux, Lavril, Lee, Liskovich, Lu, Mao, Martinet, Mihaylov, Mishra, Molybog, Nie, Poulton, Reizenstein, Rungta, Saladi, Schelten, Silva, Smith, Subramanian, Tan, Tang, Taylor, Williams, Kuan, Xu, Yan, Zarov, Zhang, Fan, Kambadur, Narang, Rodriguez, Stojnic, Edunov, and Scialom]{DBLP:journals/corr/abs-2307-09288}
Hugo Touvron, Louis Martin, Kevin Stone, Peter Albert, Amjad Almahairi, Yasmine Babaei, Nikolay Bashlykov, Soumya Batra, Prajjwal Bhargava, Shruti Bhosale, Dan Bikel, Lukas Blecher, Cristian Canton{-}Ferrer, Moya Chen, Guillem Cucurull, David Esiobu, Jude Fernandes, Jeremy Fu, Wenyin Fu, Brian Fuller, Cynthia Gao, Vedanuj Goswami, Naman Goyal, Anthony Hartshorn, Saghar Hosseini, Rui Hou, Hakan Inan, Marcin Kardas, Viktor Kerkez, Madian Khabsa, Isabel Kloumann, Artem Korenev, Punit~Singh Koura, Marie{-}Anne Lachaux, Thibaut Lavril, Jenya Lee, Diana Liskovich, Yinghai Lu, Yuning Mao, Xavier Martinet, Todor Mihaylov, Pushkar Mishra, Igor Molybog, Yixin Nie, Andrew Poulton, Jeremy Reizenstein, Rashi Rungta, Kalyan Saladi, Alan Schelten, Ruan Silva, Eric~Michael Smith, Ranjan Subramanian, Xiaoqing~Ellen Tan, Binh Tang, Ross Taylor, Adina Williams, Jian~Xiang Kuan, Puxin Xu, Zheng Yan, Iliyan Zarov, Yuchen Zhang, Angela Fan, Melanie Kambadur, Sharan Narang, Aur{\'{e}}lien Rodriguez, Robert Stojnic, Sergey Edunov,
  and Thomas Scialom.
\newblock Llama 2: Open foundation and fine-tuned chat models.
\newblock \emph{CoRR}, abs/2307.09288, 2023.

\bibitem[Tow et~al.()Tow, Bellagente, Mahan, and Riquelme]{StableLM-3B-4E1T}
Jonathan Tow, Marco Bellagente, Dakota Mahan, and Carlos Riquelme.
\newblock Stablelm 3b 4e1t.
\newblock URL \url{[https://huggingface.co/stabilityai/stablelm-3b-4e1t](https://huggingface.co/stabilityai/stablelm-3b-4e1t)}.

\bibitem[Tseng et~al.(2024)Tseng, Chee, Sun, Kuleshov, and Sa]{DBLP:conf/icml/TsengCSKS24}
Albert Tseng, Jerry Chee, Qingyao Sun, Volodymyr Kuleshov, and Christopher~De Sa.
\newblock Quip{\#}: Even better {LLM} quantization with hadamard incoherence and lattice codebooks.
\newblock In \emph{{ICML}}. OpenReview.net, 2024.

\bibitem[Wan et~al.(2024{\natexlab{a}})Wan, Wang, Liu, Alam, Zheng, Liu, Qu, Yan, Zhu, Zhang, Chowdhury, and Zhang]{DBLP:journals/tmlr/Wan0LA0LQYZZC024}
Zhongwei Wan, Xin Wang, Che Liu, Samiul Alam, Yu~Zheng, Jiachen Liu, Zhongnan Qu, Shen Yan, Yi~Zhu, Quanlu Zhang, Mosharaf Chowdhury, and Mi~Zhang.
\newblock Efficient large language models: {A} survey.
\newblock \emph{Trans. Mach. Learn. Res.}, 2024, 2024{\natexlab{a}}.

\bibitem[Wan et~al.(2024{\natexlab{b}})Wan, Wu, Zhang, Xin, Tao, Zhu, Wang, Luo, Xiong, and Zhang]{DBLP:journals/corr/abs-2406-13035}
Zhongwei Wan, Xinjian Wu, Yu~Zhang, Yi~Xin, Chaofan Tao, Zhihong Zhu, Xin Wang, Siqi Luo, Jing Xiong, and Mi~Zhang.
\newblock {D2O:} dynamic discriminative operations for efficient generative inference of large language models.
\newblock \emph{CoRR}, abs/2406.13035, 2024{\natexlab{b}}.

\bibitem[Wan et~al.(2025)Wan, Shen, Wang, Liu, Mai, and Zhang]{wan2025meda}
Zhongwei Wan, Hui Shen, Xin Wang, Che Liu, Zheda Mai, and Mi~Zhang.
\newblock Meda: Dynamic kv cache allocation for efficient multimodal long-context inference.
\newblock \emph{arXiv preprint arXiv:2502.17599}, 2025.

\bibitem[Wang et~al.(2024)Wang, Wan, Hekmati, Zong, Alam, Zhang, and Krishnamachari]{DBLP:journals/corr/abs-2401-01923}
Xin Wang, Zhongwei Wan, Arvin Hekmati, Mingyu Zong, Samiul Alam, Mi~Zhang, and Bhaskar Krishnamachari.
\newblock Iot in the era of generative {AI:} vision and challenges.
\newblock \emph{CoRR}, abs/2401.01923, 2024.

\bibitem[Xiao et~al.(2023)Xiao, Lin, Seznec, Wu, Demouth, and Han]{DBLP:conf/icml/XiaoLSWDH23}
Guangxuan Xiao, Ji~Lin, Micka{\"{e}}l Seznec, Hao Wu, Julien Demouth, and Song Han.
\newblock Smoothquant: Accurate and efficient post-training quantization for large language models.
\newblock In \emph{{ICML}}, volume 202 of \emph{Proceedings of Machine Learning Research}, pages 38087--38099. {PMLR}, 2023.

\bibitem[Xu et~al.(2024)Xu, Han, Yang, Wang, Zhu, Liu, Liu, and Che]{DBLP:conf/nips/Xu0YWZLLC24}
Yuzhuang Xu, Xu~Han, Zonghan Yang, Shuo Wang, Qingfu Zhu, Zhiyuan Liu, Weidong Liu, and Wanxiang Che.
\newblock Onebit: Towards extremely low-bit large language models.
\newblock In \emph{NeurIPS}, 2024.

\bibitem[Yuan et~al.(2023)Yuan, Shang, Song, Wu, Yan, and Sun]{DBLP:journals/corr/abs-2312-05821}
Zhihang Yuan, Yuzhang Shang, Yue Song, Qiang Wu, Yan Yan, and Guangyu Sun.
\newblock {ASVD:} activation-aware singular value decomposition for compressing large language models.
\newblock \emph{CoRR}, abs/2312.05821, 2023.

\bibitem[Yuan et~al.(2024)Yuan, Shang, and Dong]{DBLP:conf/iclr/YuanSD24}
Zhihang Yuan, Yuzhang Shang, and Zhen Dong.
\newblock {PB-LLM:} partially binarized large language models.
\newblock In \emph{{ICLR}}. OpenReview.net, 2024.

\bibitem[Zellers et~al.(2019)Zellers, Holtzman, Bisk, Farhadi, and Choi]{DBLP:conf/acl/ZellersHBFC19}
Rowan Zellers, Ari Holtzman, Yonatan Bisk, Ali Farhadi, and Yejin Choi.
\newblock Hellaswag: Can a machine really finish your sentence?
\newblock In \emph{{ACL} {(1)}}, pages 4791--4800. Association for Computational Linguistics, 2019.

\bibitem[Zhang et~al.(2022)Zhang, Roller, Goyal, Artetxe, Chen, Chen, Dewan, Diab, Li, Lin, Mihaylov, Ott, Shleifer, Shuster, Simig, Koura, Sridhar, Wang, and Zettlemoyer]{DBLP:journals/corr/abs-2205-01068}
Susan Zhang, Stephen Roller, Naman Goyal, Mikel Artetxe, Moya Chen, Shuohui Chen, Christopher Dewan, Mona~T. Diab, Xian Li, Xi~Victoria Lin, Todor Mihaylov, Myle Ott, Sam Shleifer, Kurt Shuster, Daniel Simig, Punit~Singh Koura, Anjali Sridhar, Tianlu Wang, and Luke Zettlemoyer.
\newblock {OPT:} open pre-trained transformer language models.
\newblock \emph{CoRR}, abs/2205.01068, 2022.

\bibitem[Zhao et~al.(2023)Zhao, Zhou, Li, Tang, Wang, Hou, Min, Zhang, Zhang, Dong, Du, Yang, Chen, Chen, Jiang, Ren, Li, Tang, Liu, Liu, Nie, and Wen]{DBLP:journals/corr/abs-2303-18223}
Wayne~Xin Zhao, Kun Zhou, Junyi Li, Tianyi Tang, Xiaolei Wang, Yupeng Hou, Yingqian Min, Beichen Zhang, Junjie Zhang, Zican Dong, Yifan Du, Chen Yang, Yushuo Chen, Zhipeng Chen, Jinhao Jiang, Ruiyang Ren, Yifan Li, Xinyu Tang, Zikang Liu, Peiyu Liu, Jian{-}Yun Nie, and Ji{-}Rong Wen.
\newblock A survey of large language models.
\newblock \emph{CoRR}, abs/2303.18223, 2023.

\bibitem[Zhong et~al.(2024)Zhong, Wan, Chen, Quan, and Li]{DBLP:journals/corr/abs-2406-10594}
Longguang Zhong, Fanqi Wan, Ruijun Chen, Xiaojun Quan, and Liangzhi Li.
\newblock Blockpruner: Fine-grained pruning for large language models.
\newblock \emph{CoRR}, abs/2406.10594, 2024.

\bibitem[Zhou et~al.(2024)Zhou, Ning, Hong, Fu, Xu, Li, Lou, Wang, Yuan, Li, Yan, Dai, Zhang, Dong, and Wang]{DBLP:journals/corr/abs-2404-14294}
Zixuan Zhou, Xuefei Ning, Ke~Hong, Tianyu Fu, Jiaming Xu, Shiyao Li, Yuming Lou, Luning Wang, Zhihang Yuan, Xiuhong Li, Shengen Yan, Guohao Dai, Xiao{-}Ping Zhang, Yuhan Dong, and Yu~Wang.
\newblock A survey on efficient inference for large language models.
\newblock \emph{CoRR}, abs/2404.14294, 2024.

\bibitem[Zhu et~al.(2024)Zhu, Li, Liu, Ma, and Wang]{DBLP:journals/tacl/ZhuLLMW24}
Xunyu Zhu, Jian Li, Yong Liu, Can Ma, and Weiping Wang.
\newblock A survey on model compression for large language models.
\newblock \emph{Trans. Assoc. Comput. Linguistics}, 12:\penalty0 1556--1577, 2024.

\end{thebibliography}
\bibliographystyle{plainnat}

\appendix
\newpage
\appendix
\onecolumn
\section{Appendix.}
\subsection{Pseudocode of \sysname}
\label{appendix:pseudocode}
\cref{algo:framework} shows the pseudocode of \sysname. Before compression, \sysname randomly collects a small amount of sentences as calibration data $C$, then runs the truncation-aware data whitening process as shown in~\cref{algo:whitening} to obtain the set of whitening matrix $\text{Set}_S$ for the weight to compress. After that, it runs the SVD and truncation with $\text{Set}_S$ on each weight matrix in the LLM. Instead of directly finishing the whole compression, it stores the decomposed matrices and further utilizes these matrices to run the parameter update with sequential low-rank approximation as shown in~\cref{algo:updating}.

\begin{algorithm}[ht]
\caption{Pseudocode of \sysname}
\begin{algorithmic}[1] 
\State \textbf{Input:} $M$: Original LLM
\State \textbf{Output:} $M''$: Compressed LLM by \sysname
\Procedure{\sysname}{$M$} 
    \State Randomly collect several sentences as the calibration data $C$
    \State $\text{Set}_S \gets \textproc{Truncation-Aware Data Whitening}(M, C)$
    \State $\text{Set}_W \gets M$ \Comment{Obtain the set of weights in $M$ to compress} 
    \For{$W$ \textbf{in} $\text{Set}_W$}
        \State $S \gets \text{Set}_S(W)$  \Comment{Extract the whitening matrix of current weight $W$} 
        \State $U, \Sigma, V \gets \operatorname{SVD}(WS)$   \Comment{Apply singular value decomposition on $W$}
        \State $\Sigma_1 \gets \operatorname{Trunc.}(\Sigma)$  \Comment{Truncate the smallest singular values in $\Sigma$}
        \State $W'_u \gets U(\Sigma_1)^{1/2}, W'_v \gets (\Sigma_1)^{1/2}V^TS^{-1}$ \Comment{Obtain two low-rank matrices}
        \State $M'(W) \gets W'_u, W'_v$      \Comment{Replace $W$ with $W'_u$ and $W'_v$ in $L$}
    \EndFor
    \State $M'' \gets \textproc{Parameter Update with Sequential Low-rank Approximation}(M')$
     \State \Return{$M''$}
\EndProcedure
\end{algorithmic}
\label{algo:framework}
\end{algorithm}

\begin{algorithm}[ht]
\caption{Pseudocode of Truncation-Aware Data Whitening}
\begin{algorithmic}[1] 
\State \textbf{Input:} $M$: Original LLM
\State \textbf{Input:} $C$: Calibration Data
\State \textbf{Output:} $\text{Set}_S$: Set of whitening matrices for the weight to compress in $M$
\Procedure{Truncation-Aware Data Whitening}{$M,C$} 
    \State $\text{Set}_S \gets \emptyset$  \Comment{Initialize the set of whitening matrices} 
    \State $\text{Set}_W \gets M$ \Comment{Obtain the set of weights in $M$ to compress} 
    \For{$W$ \textbf{in} $\text{Set}_W$}
        \State $X \gets M(W, C)$ \Comment{Obtain the input activation of the weight matrix $W$}
        \State $S \gets \operatorname{Cholesky\_Decomposition}(XX^T)$  \Comment{Apply cholesky decomposition on $XX^T$ }
        \State $\text{Set}_S \gets S \cup \text{Set}_S$  \Comment{Store the whitening weight matrix in the set}
    \EndFor
     \State \Return{$\text{Set}_S$}
\EndProcedure
\end{algorithmic}
\label{algo:whitening}
\end{algorithm}

\begin{algorithm}[ht]
\caption{\xin{Pseudocode of Parameter Update with Sequential Low-rank Approximation}}
\begin{algorithmic}[1] 
\State \textbf{Input:} $M'$: Compressed LLM by Truncation-aware Data Whitening
\State \textbf{Output:} $M''$: Compressed LLM with Parameter Update with Sequential Low-rank Approximation
\Procedure{Parameter Update with Sequential Low-rank Approximation}{$M'$} 
    \State $M'_u \gets \operatorname{LoRA_u}(M')$  \Comment{Fix all $W'_v$, fine-tune all $W'_u$}
    \State $M'' \gets \operatorname{LoRA_v}(M'_u)$  \Comment{Fix all $W'_u$, fine-tune all $W'_v$}
      \State \Return{$M''$}
\EndProcedure
\end{algorithmic}
\label{algo:updating}
\end{algorithm}

\subsection{Compression Loss of ASVD}
\label{subsec:ASVD_reconstruction_error}
ASVD introduces a diagonal scaling matrix $S_0$ that modifies the weight matrix to reflect the varying significance of different input channels. The linear layer is formulated as $Y=(WS_0)S_0^{-1}X$. The compression is made by keeping the largest $m$ singular value of $WS_0$:

\begin{align*}
WS_0 \approx & \sum_{i=1}^m \sigma'_i u'_i {v'}_i^T
\end{align*}
The resulting activation is expressed as:
\begin{align*}
    Y \approx & \sum_{i=1}^m \sigma'_i u'_i {v'}_i^T S_0^{-1}X ~.
\end{align*} 

The compression error $L=||(WS_0-\sum_{i=1}^{m}\sigma'_i u'_i {v'}_i^T)S_0^{-1}X||_F$ is demonstrated below:

\begin{align*}
L^2 =& ||(WS_0-\sum_{i=1}^{m}\sigma'_i u'_i {v'}_i^T)S_0^{-1}X||_F^2 \\
=& \left|\left|\sum_{i=m+1}^r \sigma'_iu'_i{v'}_i^{T}S_0^{-1}X\right|\right|_F^2 \\
=& \sum_{j=m+1}^r\sum_{i=m+1}^r \sigma'_i\sigma'_j\boldsymbol{\operatorname{Trace}}(u'_i {v'}_i^T X X^T v'_j {u'}_j^T) \\
=& \sum_{j=m+1}^r\sum_{i=m+1}^r \sigma'_i\sigma'_j\boldsymbol{\operatorname{Trace}}({u'}_j^T u'_i {v'}_i^T S_0^{-1}X X^T (S_0^{-1})^Tv'_j) \\
=& \sum_{i=m+1}^r{\sigma'}_i^2\boldsymbol{\operatorname{Trace}}({v'}_i^T S_0^{-1}X X^T (S_0^{-1})^T v'_i) \\
=& \sum_{i=m+1}^r{\sigma'}_i^2~||{v'}_i^T S_0^{-1} X||_F^2 ~,
\end{align*}

which is still a complex function that involves the activation $X$, the diagonal matrix $S_0$, the singular vector $v'_i$ and the singular value $\sigma'_i$. As a result, compression error is not directly related to the singular value, and the conventional SVD compression by truncating the smallest singular values may lead to suboptimal compression error.



\subsection{Compression Loss of \sysname}
\label{appendix:compression_loss_svdllm}
In \sysname, we also formulate the linear layer as $Y=(W S) S^{-1} X$, where $S^{-1} X X^T\left(S^{-1}\right)^T=I$. The compression is made by keeping the largest $m$ out of total $r$ singular values of $W S$. The compression loss $L$ is demonstrated as:

\begin{align*}
L^2 &=\left\|W X-W^{\prime} X\right\|_F^2=\left\|W S S^{-1} X-\boldsymbol{\operatorname{SVD}}(W S) S^{-1} X\right\|_F^2\\
& =\left\|(W S-\boldsymbol{\operatorname{SVD}}(W S)) S^{-1} X\right\|_F^2 \\
& =\left\|\left(W S-\sum_{i=1}^m \sigma_i u_i v_i^T\right) S^{-1} X\right\|_F^2 \\
& =\left\|\sum_{i=m+1}^r \sigma_i u_i v_i^T S^{-1} X\right\|_F^2 \\
& =\sum_{j=m+1}^r \sum_{i=m+1}^r \sigma_i \sigma_j \boldsymbol{\operatorname{Trace}}\left(u_i v_i^T S^{-1} X X^T\left(S^{-1}\right)^T v_j u_j^T\right) \\
& =\sum_{j=m+1}^r \sum_{i=m+1}^r \sigma_i \sigma_j \boldsymbol{\operatorname{Trace}}\left(u_i v_i^T\left(S^{-1} X X^T\left(S^{-1}\right)^T\right) v_j u_j^T\right) \\
& =\sum_{j=m+1}^r \sum_{i=m+1}^r \sigma_i \sigma_j \boldsymbol{\operatorname{Trace}}\left(u_i v_i^T v_j u_j^T\right) \\
\end{align*}
$$\because v_i^T v_i=u_i^T u_i=1 ; v_i^T v_j=u_i^T u_j=0, \boldsymbol{\operatorname{Trace}}\left(v_i v_i^T\right)=\boldsymbol{\operatorname{Trace}}\left(u_i u_i^T\right)=1, \forall i \neq j $$

$$\therefore L^2=\sum_{j=m+1}^r \sum_{i=m+1}^r \sigma_i \sigma_j \boldsymbol{\operatorname{Trace}}\left(u_i v_i^T v_j u_j^T\right)\\
=\sum_{i=m+1}^r \sigma_i^2 \boldsymbol{\operatorname{Trace}}\left(u_i v_i^T v_i u_i^T\right)=\sum_{i=m+1}^r \sigma_i^2$$

Therefore, the squared loss $L^2$ is equal to the sum of the squared singular values. Therefore, truncating the smallest singular values achieves the lowest compression loss.

\begin{figure}[t]
    \centering
    \subfigure[$W_Q\times S_Q$]{
        \includegraphics[width=1\textwidth]{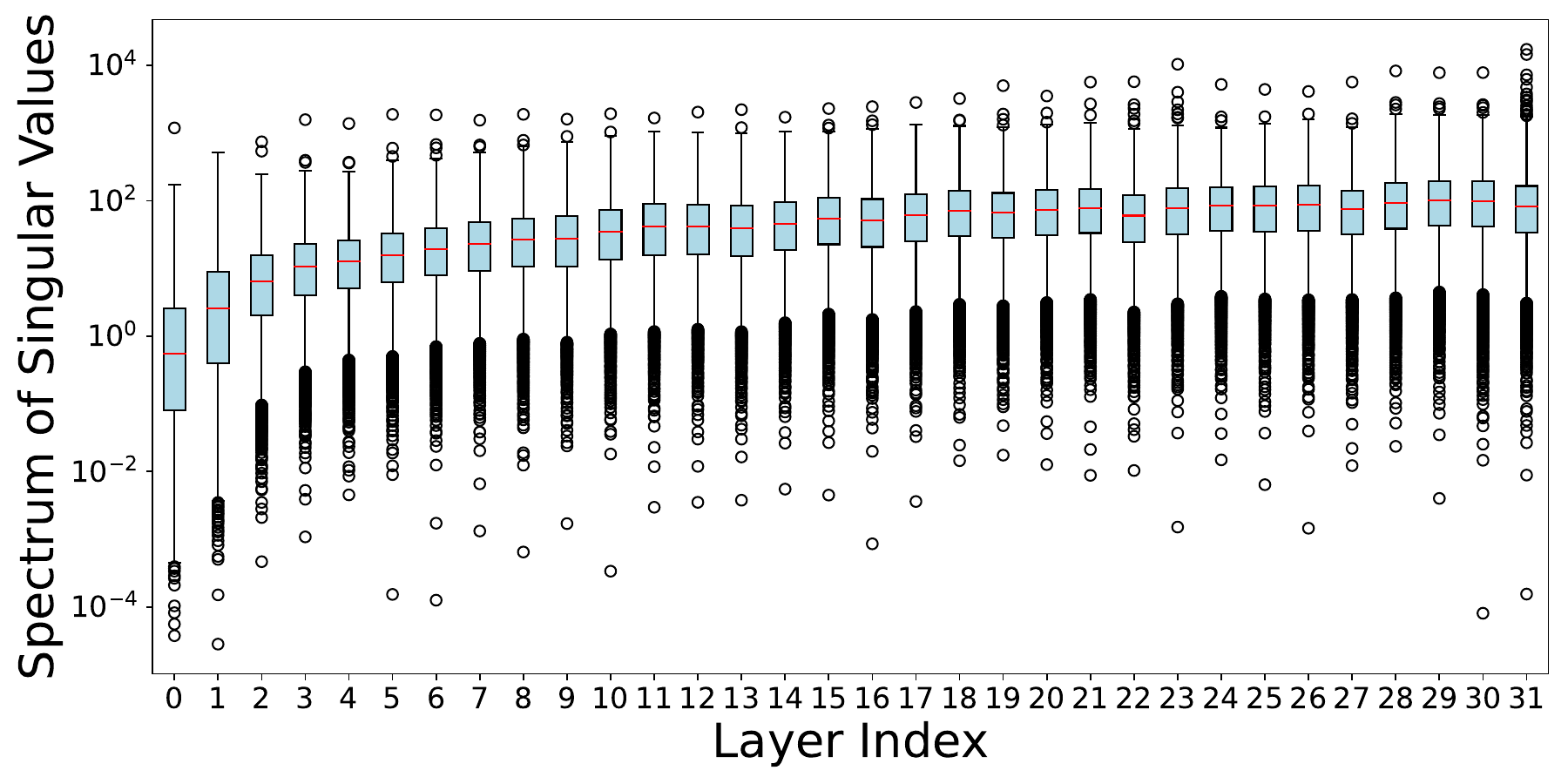}
        \label{fig:q_spectrum}
    }
    \\
    \centering
    \subfigure[$W_K\times S_K$]{
        \includegraphics[width=1\textwidth]{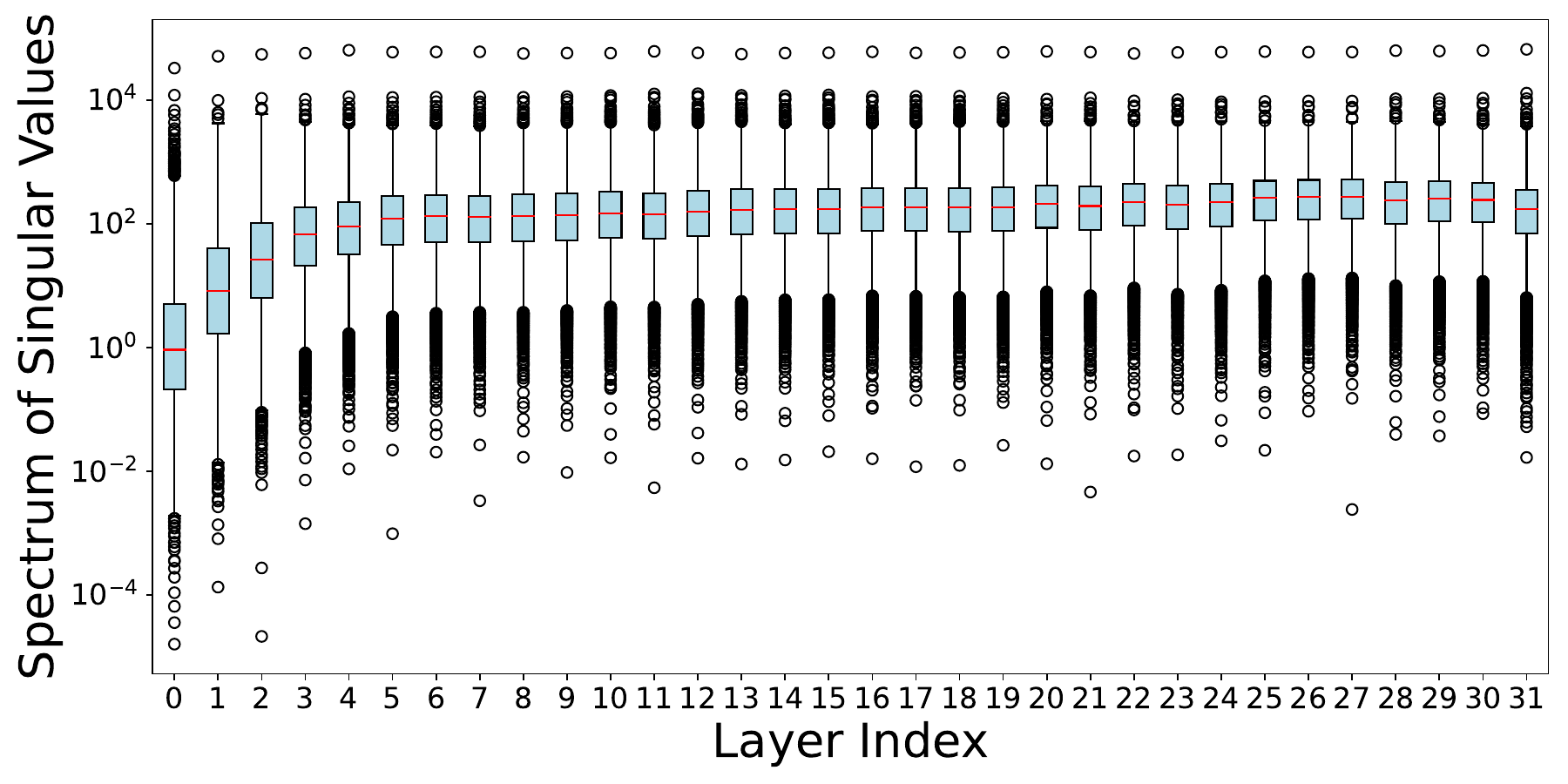}
        \label{fig:v_spectrum}
    }
    \DeclareGraphicsExtensions.
    \caption{The singular value spectrum of the decomposed matrices across layers in LLaMA-7B.}
    \label{fig:spectrum_analysis}
\end{figure}


\subsection{Spectrum Analysis of Singular Values decomposed by \sysname}
\label{appendix:spectrum_analysis}
In general, SVD is useful for compression when the matrix to be compressed shows a sharp decay of the singular values. Since \sysname decomposes the multiplication of the weight matrix $W$ and its corresponding whitening matrix $S$ instead of the original weight matrix $W$, which is different from the weight decomposition in the previous work~\citep{DBLP:journals/corr/abs-2312-05821,DBLP:conf/iclr/HsuHCLSJ22}, to study whether SVD compression is also applicable in \sysname, we select the Query ($W_Q$) and Key ($W_K$) weight matrices and show the spectrum of singular values of their multiplication with corresponding whitening matrices $S_Q$ and $S_K$. As shown in Figure~\ref{fig:spectrum_analysis}, most of the single values are less than or around 100 with only a few extremely large values, indicating that SVD is applicable in \sysname.

\subsection{Comparison with DRONE}
\label{appendix:comparison_with_drone}
\xin{
Previous work DRONE~\citep{DBLP:conf/nips/ChenYDH21} also proposes their data-aware method for SVD compression. They even provide a theoretical analysis to prove the optimal solution that their method achieves. Specifically, DRONE represents the low-rank compressed weight matrix $W'$ by $WM$. It performs SVD on both weight matrix $W = U_wS_wV_w^{T}$ and the transpose of input activation $X^T = U_xS_xV_x^{T}$ and then split these decomposed matrices as follows:}

$$
\begin{aligned}
& U_W=\left[\begin{array}{ll}
U_{W, r} & \bar{U}_{W, r}
\end{array}\right], S_W=\left[\begin{array}{cc}
S_{W, r} & 0 \\
0 & 0
\end{array}\right], V_W=\left[\begin{array}{ll}
V_{W, r} & \bar{V}_{W, r}
\end{array}\right] \\
& U_X=\left[\begin{array}{ll}
U_{X, t} & \bar{U}_{X, t}
\end{array}\right], S_X=\left[\begin{array}{cc}
S_{X, t} & 0 \\
0 & 0
\end{array}\right], V_X=\left[\begin{array}{ll}
V_{X, t} & \bar{V}_{X, t}
\end{array}\right] .
\end{aligned}
$$
\xin{
where $r$ and $k$ are the rank of the $W$ and $X$. $U_{W, r}, V_{W, r}, U_{X, t}, V_{X, t}$ denote corresponding row spaces and column spaces and $\bar{U}_{W, r}, \bar{V}_{W, r}, \bar{U}_{X, t}, \bar{V}_{X, t}$ are null spaces. Through theoretical deduction, DRONE converts the minimization of compression loss $||WX-W'X||_F = ||WX-WMX||_F$ to the minimization of $\left\|S_{W, r} V_{W, r}^T V_{X, t} S_{X, t}-S_{W, r} V_{W, r}^T M V_{X, t} S_{X, t}\right\|_F$, whose optimal value $L_{min}$ is the rank-k truncated SVD of $Z=S_{W, r} V_{W, r}^T V_{X, t} S_{X, t}$ by the fundamental property of SVD decomposition. To achieve the optimal value, DRONE formulates a solution $M = V_{W, r} S_{W, r}^{-1} Z_k S_{X, t}^{-1} V_{X, t}^T$, where $Z_k$ is the rank-k SVD truncation of $Z$. }

\xin{In short, compared with DRONE, \sysname is also optimal with the same theoretical compression loss as DRONE. Moreover, \sysname has \textbf{three} key advantages. Below is our detailed explanation.}

\xin{\textbf{\sysname is also optimal with the same theoretical compression loss as DRONE.}
The theoretical minimum compression loss $L_{min}$ is the F-norm loss of rank-k SVD truncation of $WX$, which has also been achieved by DRONE in their paper. Unlike DRONE, \sysname constructs the whitening matrix $S$ so that $S^{-1}X$ is orthonormal. Therefore, we have $||AS^{-1}X||_F = ||A||_F$. Suppose that we decompose $S$ with SVD to $U_s,S_s,V_s$, we can have $S_s = S_x$, $U_s = U_x, Us = Ux, V_s = QV_x$, where $Q$ is an orthogonal matrix. The matrix $WS$ to which \sysname applies SVD could be represented by $U_wS_wV_w^{T}U_sS_sV_s^{T}$. Suppose that we use $\boldsymbol{\operatorname{Trunc.}}(C)$ to represent the rank-k truncation of the matrix $C$ during SVD compression, the compression loss $L$ is derived as follows:}

\vspace{-5mm}
{\small
\begin{align*}
L &= ||WX-W'X||_F = ||(WSS^{-1}X -\boldsymbol{\operatorname{SVD}}(WS)S^{-1}X)||_F  = ||(WS -SVD(WS))S^{-1}X||_F\\
& = ||\boldsymbol{\operatorname{Trunc.}}(WS)S^{-1}X||_F = ||\boldsymbol{\operatorname{Trunc.}}(WS)||_F\\
&=||\boldsymbol{\operatorname{Trunc.}}(U_wS_wV_w^{T}U_sS_sV_s^{T})||_F =||\boldsymbol{\operatorname{Trunc.}}(WXQ^T)||_F = L_{min}
\end{align*}
}

\xin{Therefore, \sysname shares the same theoretical compression loss as DRONE. }

\xin{\textbf{Advantage \#1: DRONE incurs out-of-memory when compressing LLMs due to its requirement of storing the full large-scale activations, whereas \sysname is feasible.} To achieve data-awareness during compression, DRONE caches all input activations $X$ and spans them to calculate the corresponding singular vectors and singular values. In the DRONE paper, the authors apply DRONE to small LMs such as BERT. However, the activations generated by LLMs are often extremely large and are much larger than BERT. 
For example, to compress LLaMA-7B with 5,000 calibration data by DRONE, the total memory to cache the activation for a single weight matrix at a time is 5000 (data number) $\times$ 256 (seq\_len) $\times$ 11008 (dim) $\times$ 32 (fp32) $\div$ 1024 $\div$ 1024 $\div$ 1024 $=$ 419GB, which is more than 5 times larger than the memory provided by the NVIDIA A100 GPU, which has 80GB memory. Therefore, applying DRONE for LLM compression is infeasbile.  }

\dronelossTable

\xin{In contrast, \sysname incrementally updates its $XX^T$ matrix by adding the $xx^T$ of each new input $x$. As such, \sysname eliminates the need to store the full activations, which requires only the storage of the matrix $XX^T$, which is considerably smaller than the full input activation. To compress LLaMA-7B with 5,000 calibration data, \sysname requires only 11008 $\times$ 11008 $\times$ 32 $\div$ 1024 $\div$ 1024 $\div$ 1024 $=$ 3.6GB. Compared to DRONE, \sysname achieves 116.38 times less memory reduction than DRONE. In our paper, we use WikiText-2 as a dataset. If we use DRONE, the total memory to cache the activation of a single weight matrix is larger than 24,600GB. In contrast, \sysname still requires 3.6GB of memory, which is more than 6,000 times less than DRONE. Due to this advantage, \sysname is much more practical to compress LLMs of size 7B or larger compared to DRONE.}

 \dronetimeTable
\xin{\textbf{Advantage \#2: \sysname incurs much shorter compression time compared to DRONE.} DRONE involves more complex matrix operations, leading to longer compression time compared to \sysname. To illustrate this, we measured the time required by DRONE and \sysname to compress randomly generated weight and activation matrices of varying shapes under 50\% compression ratio. The results show that \sysname is approximately three times faster than DRONE.}

\xin{\textbf{Advantage \#3: \sysname has better numerical stability, which leads to superior empirical compression loss.}
 While \sysname shares the same theoretical compression loss as DRONE, DRONE’s higher complexity—stemming from additional SVD operations and inverse calculations on large-scale matrices—makes it less numerically stable compared to \sysname. This often results in higher empirical compression losses. To illustrate this, we compare \sysname and DRONE in terms of empirical compression losses for randomly generated matrices of various shapes. We also include the theoretical minimum value, represented by the rank-k Frobenius norm loss of $WX$. The results are summarized in the following table. As shown, we observe that \sysname achieves lower empirical compression losses than DRONE, underscoring its superior numerical stability.}

\flapTable

\subsection{Comparison with FLAP}
\label{appendix:flap_comparison}
\xin{
Recent work FLAP~\citep{DBLP:conf/aaai/AnZYTW24} is also a post-training structured-pruning method. Below we compare the perplexity of \sysname and FLAP on WikiText-2 under different compression ratios when compressing LLaMA-7B. As shown in~\cref{tab:flap}, \sysname consistently outperforms FLAP, especially under high compression ratios. }

\scalingTable

\subsection{Comparison with smaller LLMs pre-trained from scratch}
\label{appendix:pretrained_comparison}
To compare the performance between \sysname and scratch training, following the previous experimental design~\citep{DBLP:conf/nips/MaFW23}, we compress LLaMA-7B to the size of the 3B parameter with \sysname and select StableLM-3B~\citep{StableLM-3B-4E1T} as the baseline for comparison. As shown in Table~\ref{tab:scaling}. LLaMA-3B compressed from LLaMA-7B by \sysname achieves better accuracy in all datasets, indicating that \sysname could even achieve better accuracy than some scratch training methods. Furthermore, \sysname ensures higher throughput and lower memory consumption than StableLM-3B as shown in the table, which also meets our efficiency analysis and discussion in Section~\ref{subsec:inference_speedup}. 


\timeTable

\subsection{Compression Speed Evaluation}
\label{appendix:compression_speed}
In addition to compression performance, we also evaluate the compression speed of \sysname and the baselines. Specifically, we measured the GPU hours used for \sysname and ASVD when compressing LLaMA-7B under 20\% compression ratio on an A100 GPU. The results are shown in~\cref{tab:time}. As shown, ASVD takes about $5.5$ hours, while \sysname completes the compression process in $3.5$ hours, which is $36\%$ times faster. When breaking down the time, most of the time consumed by ASVD is dedicated to searching for the specific compression ratio for each weight matrix based on its calculated importance score. In contrast, \sysname maintains a consistent compression ratio in all weight matrices and thus eliminates the time-consuming search process.

\subsection{Contents Generated from LLMs Compressed by \sysname and ASVD}
\label{sec:appendix_example}
Some examples of sentences generated by LLaMA-7B compressed with \sysname and ASVD are shown in~\cref{tab:visualization_lora}. 
The sentences generated by the model compressed by \sysname exhibit better fluency, relevance, and informativeness compared to those compressed by ASVD. More importantly, when the compression ratio is increased to 40\%, the previous state-of-the-art method ASVD completely loses its generation ability. 
In contrast, even when the compression ratio is up to 80\%, \sysname is still capable of generating complete sentences.
\exampleLoraTable




\subsection{More Ablation Studies}
\label{appendix:more_ablation}
\xin{
\textbf{\sysname + Normal LoRA Fine-tuning v.s. \sysname + Sequential LoRA fine-tuning.} To illustrate the superiority of the designed parameter update with the sequential low-rank approximation in \sysname, which is a kind of sequential LoRA fine-tuning strategy over the normal LoRA fine-tuning strategy, we compare the compression performance of \sysname by applying either of these two fine-tuning strategies. Let's denote \sysname (SFT) as \sysname by applying sequential LoRA fine-tuning and \sysname (NFT) as \sysname by applying normal LoRA fine-tuning. As shown in~\cref{tab:lora_add}, \sysname (SFT) consistently outperforms \sysname (NFT), which also reaffirms our analysis in~\cref{subsec:lora_finetunig} that optimizing both low-rank matrices $W_u, W_v$ at the same time is not stable and may lead to poor fine-tuning performance.}

\loraaddTable

\xin{
\textbf{ASVD + Sequential LoRA Fine-tuning v.s. \sysname + Sequential LoRA Fine-tuning.} Although the designed sequential LoRA fine-tuning strategy could also be applied in other SVD-based LLM compression methods, other methods' performance is still poorer than \sysname even when integrated with this strategy for enhancement. To illustrate this, we compare the performance of the previous state-of-the-art method ASVD when applied with the sequential LoRA finetuning with \sysname. Let's denote \sysname (SFT) as \sysname by applying sequential LoRA fine-tuning and ASVD (SFT) as ASVD by applying sequential LoRA fine-tuning. As shown in~\cref{tab:lora_add}, \sysname (SFT) consistently outperforms ASVD (SFT) under various compression ratios.}


\subsection{Limitations}
There are three limitations of \sysname, which are left for future work.

\textbf{(1) The compression accuracy still needs to be improved under the high compression ratio.}
Although \sysname has achieved the state-of-the-art performance compared to previous works such as FWSVD and ASVD, its compression accuracy still suffers from degradation, especially under the high compression ratio. To enhance the practicability of \sysname in real-world scenario, its accuracy should be at least comparable to that of the quantization method, including both low-bit and high-bit quantization, rather than being combined with the quantization methods for usage.

\textbf{(2) The latency of \sysname needs to be optimized while being used to compress the KV cache.} As discussed in~\cref{subsec:inference_speedup}, \sysname can also be applied to compress the KV cache for memory saving during inference. However, this benefit does not come free. In fact, due to the additional calculation caused by recovering the original key and value states, the inference speed will be impacted by compressing the KV cache with \sysname, which should be optimized in the future.

\textbf{(3) \sysname should be better guided for high-quality generation.} Although \sysname achieves low perplexity, as demonstrated in~\cref{tab:dataset_acc}, it is still possible to generate low-quality content, such as repeated words even under low compression ratios. This phenomenon of compressed LLM has also been mentioned in previous work~\citep{DBLP:conf/nips/MaFW23} and should be eliminated in the future by guiding the compressed LLM for outputting high-quality generation.

\end{document}